\documentclass{article}

\PassOptionsToPackage{table}{xcolor}

\usepackage{times}
\usepackage{soul}
\usepackage{url}
\usepackage{makecell}
\usepackage{hyperref}
\usepackage{graphicx}
\usepackage[small]{caption}
\usepackage{amsmath}
\usepackage{amsthm}
\usepackage[switch]{lineno}
\usepackage{longtable}
\usepackage{supertabular}
\usepackage{multirow}
\usepackage{tikz}                  
\usepackage[edges]{forest}
\usepackage{natbib}

\usepackage[in]{fullpage}  
\usepackage{arxiv} 
\usepackage{mathpazo} 

\usepackage[utf8]{inputenc} 
\usepackage[T1]{fontenc}    
\usepackage{url}            
\usepackage{booktabs}       
\usepackage{amsfonts}       
\usepackage{microtype}      
\usepackage{lipsum}         

\usepackage{graphicx}  
\usepackage{float} 
\usepackage{subfigure} 
\usepackage{algorithm}
\usepackage{tabulary}

\usepackage{amssymb}
\makeatletter

\newcommand{\Rmnum}[1]{\expandafter\@slowromancap\romannumeral #1@}
\makeatother

\usepackage{pifont}
\usepackage{xcolor}

\hypersetup{
    colorlinks=true,
    linkcolor=green,
    citecolor=red
}
\usepackage{tcolorbox}

\usepackage{natbib}
\usepackage{microtype} \usepackage{graphicx} \usepackage{booktabs} \usepackage{amsmath,amssymb,mathtools,amsthm,bm} \usepackage{xspace} \usepackage{xcolor} \usepackage{colortbl} \usepackage{framed} \usepackage{tcolorbox} \usepackage{adjustbox} \usepackage{multirow} \usepackage{makecell} \usepackage{threeparttable} \usepackage{pifont} \usepackage{wrapfig} \usepackage[normalem]{ulem} \usepackage{array} \usepackage{tocvsec2} \usepackage{xpatch} \usepackage{inconsolata} \usepackage{textcomp} \usepackage[utf8]{inputenc} 
\usepackage[T1]{fontenc} 
\usepackage{hyperref} 
\usepackage{url} 
\usepackage{amsfonts} 
\usepackage{nicefrac} 
\usepackage{algorithm} 
\usepackage{algpseudocode} 
\usepackage{tabularx} 
\usepackage{float}

\hypersetup{
    colorlinks=true,
    linkcolor=blue,
    citecolor=orange,
    urlcolor=blue
}

\definecolor{red}{rgb}{1, 0, 0}

\definecolor{green}{rgb}{0, 1, 0}

\definecolor{blue}{rgb}{0, 0, 1}

\definecolor{mydarkblue}{rgb}{0, 0.08, 0.45}
\definecolor{rowblue}{HTML}{EAF2FF}
\definecolor{accA}{HTML}{FF8C2D}
\definecolor{accB}{HTML}{00A8C6}
\definecolor{lightblue}{RGB}{100,150,255}

\newcolumntype{Y}{>{\centering\arraybackslash}p{1.35cm}}
\newcolumntype{M}[1]{>{\centering\arraybackslash}p{#1}}
\newcolumntype{L}[1]{>{\raggedright\arraybackslash}p{#1}}

\tcbuselibrary{listings,breakable,skins}
\usepackage{fvextra}
\graphicspath{{./}{figures/}}
\usepackage{pifont}
\newcommand{\ToMap}{\textsc{ToMap}\xspace}

\newcommand{\D}{\mathcal{D}}
\newcommand{\F}{\mathcal{F}}
\newcommand{\Pvr}{\mathcal{P}}

\newcommand{\best}[1]{\textbf{#1}}

\newcommand{\gain}[1]{\textcolor{red}{\scriptsize\,(\ensuremath{\uparrow}#1)}}
\newcommand{\maxcost}[1]{#1$^{\dagger}$}

\title{Efficient Test-Time Optimization for Multi-Agent Proof Autoformalization}

\usepackage{authblk}
\author{%
   Tian-Shuo Liu\textsuperscript{\rm 1,2,*}, Shiyuan Zhang\textsuperscript{\rm 1,*}, Zijie Geng\textsuperscript{\rm 3}, Haoyu Liu\textsuperscript{\rm 1}, Runjie Xu\textsuperscript{\rm 1}, Pengyuan Wang\textsuperscript{\rm 1}, Lei Yuan\textsuperscript{\rm 1}, Yang Yu\textsuperscript{\rm 1,2,$\diamond$}\\
  \textsuperscript{\rm 1} National Key Laboratory for Novel Software Technology, Nanjing University, Nanjing, China \\\;\&\;  School of Artificial Intelligence, Nanjing University, Nanjing, China\\
  \textsuperscript{\rm 2}Polixir Technologies, Nanjing, China \\
  \textsuperscript{\rm 3} University of Science and Technology of China, Hefei, China\\

  \textsuperscript{*} Equal contribution\\
  \textsuperscript{$\diamond$} Corresponding: yuy@nju.edu.cn
}

\date{}
\begin{document}

\maketitle

\begin{abstract}
Full-proof autoformalization bridges extensive mathematical proofs in natural language with  formally validated reasoning, offering a pathway to elevate the ceiling of verifiable mathematical reasoning. Unlike statement-level formalization, proof autoformalization is a long-horizon challenge requiring coordination of claims, contexts, and dependencies across many proof steps, yet has only recently come under focused study. Current approaches either rely on costly model training or apply excessive, unguided repair at inference time.
To this end, we introduce \textsc{ToMap}, a multi-agent framework that structures proof autoformalization as a Decomposer-Formalizer-Prover pipeline with efficient test-time optimization guided by formal verification and semantic rubrics for proof quality. Rather than distributing test-time compute across all agents, we perform bottleneck analysis and identify the Decomposer as the critical bottleneck: the quality of its atomic, self-contained proof units directly determines whether downstream agents can successfully formalize and prove each step. \textsc{ToMap} therefore treats the Formalizer and Prover as downstream executors and efficiently focuses test-time compute on Decomposer refinement. This refinement follows a loop inspired by GEPA, evolving prompts over candidate decompositions and using formal verification progress together with semantic proof rubrics to define a Pareto frontier that guides the next decomposition update.
Experiments on \textsc{ProofFlowBench} show that \textsc{ToMap} improves over the best previous method by 19.0\% when evaluated by both syntactic correctness and semantic faithfulness, while requiring lower test-time cost. Scaling analysis shows that most gains emerge within a few iterations of decomposition evolution, guiding test-time budget selection.
\end{abstract}

\section{Introduction}

Large language models have made rapid progress in mathematical reasoning~\citep{llmsurveymath, lewkowycz2022solving, azerbayev2023llemma}. Among these advances, mathematical proof remains a particularly stringent target, as plausible solutions do not guarantee that intermediate claims, dependencies, and side conditions are logically valid. This has motivated a growing body of work on combining informal reasoning with formal verification, where proof assistants provide machine-checkable guarantees~\citep{dekoninck2025open, jiang2022draft, ren2025deepseek, chen2025seed}. Autoformalization, translating informal mathematics into formal languages, is the central interface in this effort~\citep{wu2022autoformalization}. Early work focused on statement-level formalization. More recently, full-proof autoformalization has gained increasing attention, as translating complete proofs more effectively bridges informal and formal mathematics by leveraging the abundance of natural-language proofs while retaining the rigorous verifiability of formal systems.

This shift, however, fundamentally changes the autoformalization problem. A theorem statement is a localized translation target~\citep{wu2022autoformalization}, whereas a full proof is a long-horizon construction of intermediate claims, local assumptions, and dependencies that must compose into a globally checkable whole~\citep{jiang2022draft}. Different stages of this translation demand distinct LLM capabilities acquired during post-training: natural-language reasoning for decomposition, formal-language fluency for formalization, and tactic generation for proof search under verifier feedback~\citep{dong2024prod, ren2025deepseek, baba2025prover,liu2026offpolicy}. This naturally leads to agentic, verifier-in-the-loop pipelines in which specialized agents handle each stage while a proof assistant provides executable feedback~\citep{cabral2025proofflow, m2f, monotonic}. Proof-level paired supervision remains scarce and costly~\citep{jana2026proofbridge, chen2025seed}, making large-scale training difficult. Recent systems therefore rely on Lean as a proof assistant at inference time, using staged repair or iterative refinement to improve candidate proofs~\citep{cabral2025proofflow, m2f, monotonic}. In full-proof settings, however, verifier feedback arrives only after a long generation chain and rarely reveals which upstream decision caused the failure or how the next attempt should differ. This makes efficient test-time optimization a central challenge: under a fixed budget, where should feedback be directed in the pipeline, and how can it drive targeted improvement rather than unguided retries?

Following recent agentic approaches to proof generation~\citep{cabral2025proofflow, ren2025deepseek, dong2024prod}, we decompose full-proof autoformalization into three specialized agents: a Decomposer that splits the informal proof into sub-propositions, a Formalizer that translates each into Lean, and a Prover that closes each goal. To perform efficient test-time optimization in this pipeline, our analysis first identifies the Decomposer as the highest-leverage stage: when it produces atomic, self-contained proof units with explicit dependencies, the downstream agents face only local obligations that are already within their capabilities. This aligns with prior work that constructs sketches or subgoals before detailed
proving~\citep{jiang2022draft, cabral2025proofflow, ren2025deepseek, dong2024prod}, demonstrating that the key difficulty is constructing the right decomposition.

Building on this diagnosis, we introduce Test-time Optimization for Multi-Agent Proof autoformalization (\textsc{ToMap}), an efficient test-time optimizer for the decomposition agent. 
For each input, \ToMap maintains a pool of candidate decompositions that split the informal proof into sub-propositions. Inspired by reflective prompt evolution~\citep{2025gepa}, \ToMap iteratively evolves these decompositions, evaluating each candidate against a fixed set of proof-quality rubrics that score semantic faithfulness, atomicity, self-containedness, and dependency consistency. The rubric scores, computed before any downstream formalization or proving, define a Pareto frontier that guides which candidates are retained and revised across iterations. Only when the rubrics judge a candidate ready does \textsc{ToMap} commit it to the downstream Formalizer–Prover executor for Lean verification. This two-level design is central to efficiency: rubric feedback is cheap and dense, steering the bulk of the search, while Lean verification provides faithful but expensive ground-truth signals reserved for promising candidates. By balancing cheap proxy feedback against costly true verification, \textsc{ToMap} converts limited test-time compute into directed decomposition improvement rather than unguided pipeline retries. Experiments on \textsc{ProofFlowBench} and miniF2F demonstrate that \textsc{ToMap} outperforms both training-based and training-free baselines, achieving higher Lean pass rates and stronger semantic faithfulness at lower time cost. Ablations across evolution epochs reveal a clear time–performance trade-off that provides practical guidance for iteration budget selection.

\section{Full-Proof Autoformalization as a Multi-Agent LLM System}
\label{sec:formulation}

\subsection{Problem Setup}

Let $x=(t,p)$ denote an input consisting of a natural-language theorem statement $t$ and an informal proof $p$.
Full-proof autoformalization aims to produce a Lean proof file $y$ that is accepted by the verifier and faithfully follows the reasoning in $p$, matching the full-proof setting studied in prior work~\citep{jana2026proofbridge}.
A monolithic whole-proof generator samples
\begin{equation}
    y \sim M(\cdot \mid t,p),
\end{equation}
and receives feedback from the Lean verifier $\mathcal{V}$. Unlike statement-level formalization, full-proof translation requires recovering implicit assumptions, dependency chains, and side conditions from informal text, expressing each local claim in formal language, and generating valid tactics for every goal. These span distinct capabilities: natural-language proof understanding, formal-language fluency, and tactic-level proof search. We therefore model full-proof autoformalization as a multi-stage LLM system whose intermediate outputs can be inspected and revised.

The goal is to maximize end-to-end verification success under a fixed test-time budget $B$, measured as wall-clock time in a fixed evaluation environment $\mathcal{H}$, including the hardware, Lean version, model endpoints, maximum parallelism, and scheduling policy:
\begin{equation}
    \theta^* = \arg\max_{\theta}\;\mathbb{E}_{x \sim \mathcal{D}}
    \bigl[\mathbf{1}[r_{\mathrm{full}}(\tau(x;\theta)) = \texttt{pass}]\bigr]
    \quad \text{s.t.} \quad \mathrm{Cost}(\theta, B) \leq B.
    \label{eq:objective}
\end{equation}

\subsection{Multi-Agent Proof Autoformalization Pipeline}

\begin{figure}[t]
\centering
\includegraphics[width=0.82\linewidth]{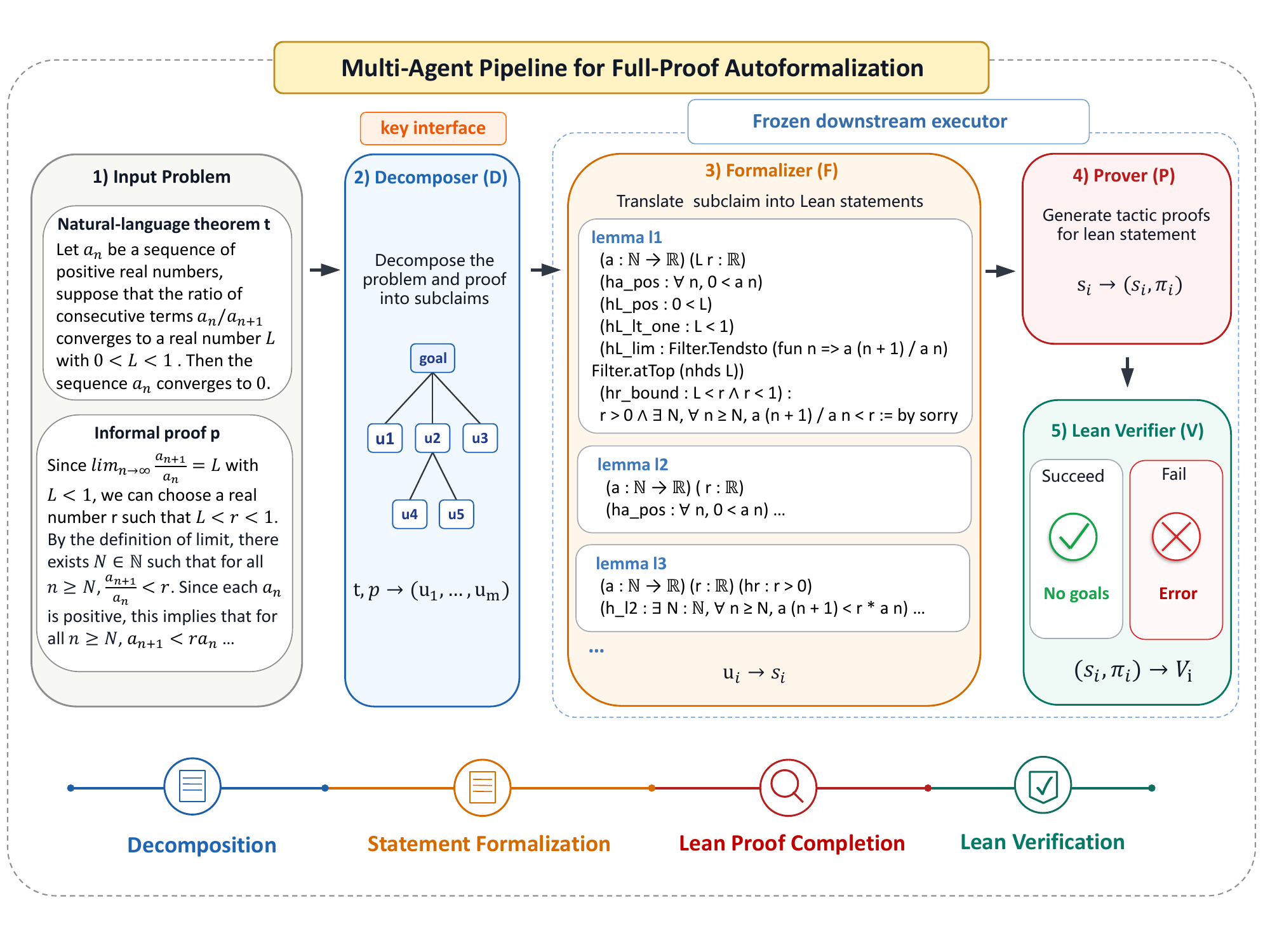}
\vspace{-2mm}
\caption{\textbf{Full-proof autoformalization as a multi-agent pipeline.} A natural-language theorem and informal proof are first decomposed into proof units with local context. The downstream executor then formalizes each unit, proves each local obligation, and verifies them.}
\vspace{-2mm}
\label{fig:pipeline}
\end{figure}

Figure~\ref{fig:pipeline} illustrates the overall system.
We write the LLM agent pipeline as
\begin{equation}
    \Psi_{\theta}
    =
    \Pvr_{\theta_P}
    \circ
    \F_{\theta_F}
    \circ
    \D_{\theta_D},
    \label{eq:agent_pipeline}
\end{equation}
where $\D$ is the natural-language Decomposer, $\F$ is the Formalizer, and $\Pvr$ is the Prover.
These are the three LLM-driven agents in our system.
The prompt or instruction parameters are
\begin{equation}
    \theta=(\theta_D,\theta_F,\theta_P).
\end{equation}
The Lean verifier $\mathcal{V}$ is not part of the agent pipeline; it is an external checker that evaluates the generated Lean proof file and returns feedback.
\ToMap adapts $\theta_D$ at test time while keeping $\theta_F$ and $\theta_P$ fixed.

Given an input $x=(t,p)$, the Decomposer first produces a proof-unit decomposition
\begin{equation}
    z=\D_{\theta_D}(x)=(\mathcal{G}_z,\{n_1,\ldots,n_m\}),
\end{equation}
where $\mathcal{G}_z$ records dependencies among proof units.
Conceptually, each proof unit is represented as
\begin{equation}
    n_i=(q_i,u_i,a_i,d_i,\kappa_i),
\end{equation}
where $q_i$ is a local claim, $u_i$ are in-scope variables, $a_i$ are assumptions, $d_i$ are dependencies on earlier units, and $\kappa_i$ contains side conditions or proof hints.
This tuple is a modeling abstraction; in implementation, each $n_i$ is written as natural-language text.

The purpose of decomposition is to turn a long informal proof into local obligations that downstream agents can plausibly formalize and prove.
A useful proof unit should be atomic enough for the Formalizer and Prover to handle, but complete enough to specify the right claim, assumptions, side conditions, and dependencies.
Decomposition quality is therefore not the same as granularity alone: simply splitting a proof into more units is not necessarily better.

For each proof unit $n_i$, the system constructs a local context $c_i$ from the theorem, the ordered decomposition, and the relevant earlier units. 
The Formalizer produces a formal statement $s_i=\F_{\theta_F}(n_i,c_i)$, and the Prover produces a proof script $\pi_i=\Pvr_{\theta_P}(s_i,c_i)$. 
The statement-proof pairs are assembled into a candidate Lean proof file:
\begin{equation}
    y=\operatorname{Assemble}\!\left(x,z,\{(s_i,\pi_i)\}_{i=1}^{m}\right),
    \qquad
    r=\mathcal{V}(y).
\end{equation}
Here $r$ denotes the verifier feedback, including the full-proof result and any available local errors. 
Later sections use this feedback to analyze bottlenecks and guide test-time optimization.

\section{Weak-Link Analysis of the Multi-Agent Pipeline}
\label{sec:weaklink}

Before allocating test-time computation, we first ask where verifier feedback is most useful.
In a modular proof-generation system, the stage at which a Lean error is observed is not necessarily the stage whose revision would most improve the final proof.
Following the bottleneck-localization and controlled-intervention principles distilled from recent weak-link and causal analyses of multi-agent systems~\citep{weak-link,causal_analysis}, we measure the \emph{feedback recoverability} of each stage: when the same Lean diagnostic trace is exposed to only one editable stage while the remaining stages are held fixed, which stage yields the largest improvement in end-to-end verification?

\paragraph{Protocol.}
For each input $x=(t,p)$, we first execute the unmodified Decomposer-Formalizer-Prover pipeline and record a structured Lean trace,
\begin{equation}
\tau(x;\theta)=
\bigl(
x,\,
z,\,
\{(n_i,c_i,s_i,\pi_i,r_i)\}_{i=1}^{m},\,
r_{\mathrm{full}}
\bigr),
\end{equation}
where $r_i$ contains obligation-level verifier diagnostics, including elaboration outcomes, goal states, and Lean error messages, and $r_{\mathrm{full}}$ denotes whether the assembled proof file is accepted by Lean.
We then construct three matched revision conditions.
Each condition receives the same diagnostic trace, uses the same verification procedure, and is given the same correction budget; the only varying factor is the stage whose output may be revised.

\begin{itemize}
    \item \textbf{D-Intervene.}
    Only the Decomposer output is editable.
    Conditioned on the verifier trace, the Decomposer revises the natural-language proof-unit specification, including local claims, assumptions, side conditions, and context requirements.
    The Formalizer and Prover are kept fixed and are re-executed on the revised decomposition.

    \item \textbf{F-Intervene.}
    The decomposition and local contexts are fixed.
    Conditioned on the same verifier trace, the Formalizer revises only the formal statements $\{s_i\}_{i=1}^{m}$.
    The Prover is kept fixed and is re-executed on the revised statements.

    \item \textbf{P-Intervene.}
    The decomposition, local contexts, and formal statements are fixed.
    Conditioned on the same verifier trace, the Prover revises only the proof scripts $\{\pi_i\}_{i=1}^{m}$.
\end{itemize}

This matched design isolates the revision locus under identical verifier feedback, rather than comparing generic self-correction prompts.
The resulting gain estimates how much end-to-end verification can be recovered by correcting each stage under the same diagnostic information and compute budget.

\paragraph{Results.}
\begin{wrapfigure}{r}{0.45\textwidth}
    \centering
    \vspace{-2.5em}
    \includegraphics[width=0.45\textwidth]{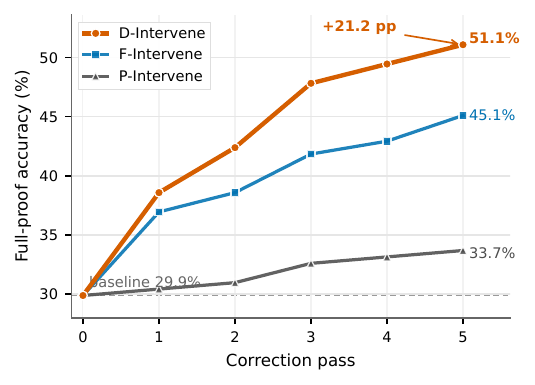}
    \caption{Full-proof accuracy over five correction passes for each interface intervention.}
    \label{fig:weaklink_intervention}
    \vspace{-1.8em}
\end{wrapfigure}
Figure~\ref{fig:weaklink_intervention} shows full-proof accuracy over five correction passes on all 184 benchmark samples; samples without a successful baseline trace are counted as failures.
D-Intervene consistently achieves the highest accuracy, reaching $94/184 = 51.1\%$ at pass~5, compared with $83/184 = 45.1\%$ for F-Intervene and $62/184 = 33.7\%$ for P-Intervene.
Under a fixed verifier-feedback budget, revising the natural-language proof-unit specification yields the largest gain in end-to-end verification.

This ordering reveals why the Decomposer is the right target for \ToMap. Intuitively, the Decomposer sets the difficulty of every downstream obligation. Well-scoped, atomic proof units reduce formalization and proving to local tasks while poorly specified units leave the executor with ambiguous or incomplete goals. That D-Intervene achieves the largest gains despite a fixed 32B executor confirms that the bottleneck is decomposition quality. \ToMap therefore directs all test-time compute to the Decomposer, treating the
  Formalizer and Prover as fixed executors.

\section{\ToMap: Test-Time Optimization of Proof Decompositions}

\label{sec:method}

\begin{figure}[t]
\centering
\includegraphics[width=\linewidth]{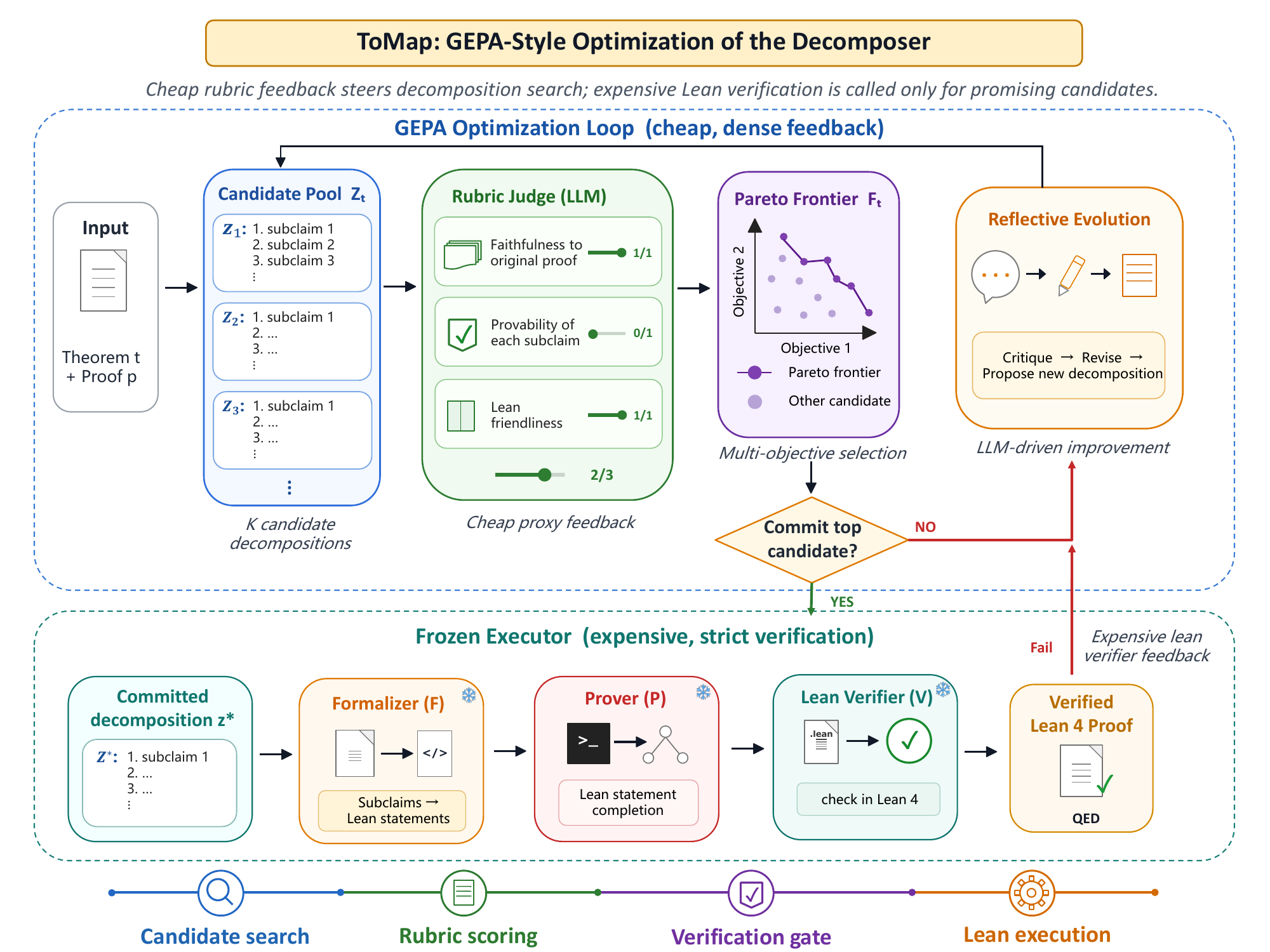}
\caption{\textbf{\textsc{ToMap}: GEPA-style test-time optimization of proof decompositions.}
\textsc{ToMap} concentrates test-time compute on the Decomposer interface while keeping the downstream Formalizer-Prover-Lean executor fixed.
It maintains a pool of candidate natural-language decompositions, scores them with proof-quality rubrics measuring semantic faithfulness, atomicity, self-containedness, and dependency consistency, and uses the resulting Pareto frontier to guide reflective evolution.
Only candidates that pass the rubric-based commit gate are sent to the expensive downstream executor for Lean verification, allowing cheap dense feedback to steer most of the search while reserving faithful verification for promising decompositions.}
\label{fig:tomap_gepa}
\end{figure}

The weak-link analysis of Section~\ref{sec:weaklink} identifies the Decomposer as the most recoverable agent in the pipeline. Under a fixed test-time budget, we therefore prioritize efficient optimization of the decomposition stage. However, simply applying self-correction through the downstream Formalizer-Prover executor with Lean verification is inefficient: each Lean call is expensive,
and a binary pass/fail signal offers little guidance for the next revision.                                               
\ToMap instead decouples search from verification.
A pool of candidate decompositions is evolved using cheap LLM-based rubric feedback, and the downstream pipeline is invoked only when a candidate meets a quality threshold.
By using dense rubric guidance for the bulk of the search and reserving selective Lean calls for promising candidates, this two-level design converts limited test-time compute into directed decomposition improvement.

\paragraph{Rubric scoring.}
Each candidate decomposition $z$ is evaluated by a three-dimensional rubric vector
\begin{equation}
    \rho(z) = \bigl(\rho_{\mathrm{faithful}}(z),\;\rho_{\mathrm{provable}}(z),\;\rho_{\mathrm{Lean friendly}}(z)\bigr) \in [0,1]^3,
\end{equation}
where $\rho_{\mathrm{faithful}}$ measures semantic faithfulness to the original proof $p$, $\rho_{\mathrm{provable}}$ measures the provability of each proof unit, and $\rho_{\mathrm{Lean friendly}}$ measures whether a proof unit is Lean-friendly, such as by eliminating ambiguous natural-language descriptions.
For compact notation, let
$
    \mathcal{R}
    =
    \{\mathrm{faithful}, \mathrm{provable}, \mathrm{Lean friendly}\}$.
Each dimension is scored by an LLM judge conditioned on $x$ and $z$, so that $\rho_j(z)$ denotes the rubric score of candidate $z$ on dimension $j \in \mathcal{R}$. 
The rubric vector provides dense, low-cost feedback that approximates downstream verifiability without executing the full pipeline.

\paragraph{Pareto-guided evolution.}
\ToMap maintains a candidate pool $\mathcal{Z}_t$ of size $K$.
Following GEPA~\citep{2025gepa}, candidates on the Pareto frontier
\begin{equation}
    \mathcal{F}_t = \bigl\{z \in \mathcal{Z}_t \;\big|\; \nexists\, z' \in \mathcal{Z}_t \text{ s.t. } \rho(z') \succ \rho(z)\bigr\},
\end{equation}
where $\succ$ denotes component-wise dominance.
Candidates in $\mathcal{F}_t$ serve as parents for the next generation.
At each iteration, a reflective LLM proposer selects a parent from $\mathcal{F}_t$, receives a structured critique derived from its rubric scores and the original proof, and generates a revised decomposition.
The pool is updated by admitting the new candidate and discarding dominated ones.
Maintaining a Pareto frontier rather than a scalar objective ensures that evolutionary pressure is distributed across all quality dimensions.

\paragraph{Verification gating.}
A candidate $z$ is committed to the Formalizer-Prover executor for Lean verification only when
\begin{equation}
    \min_{j \in \mathcal{R}} \rho_j(z) \geq \epsilon .
\end{equation}
Candidates that pass the gate are assembled into a Lean proof file, checked by $\mathcal{V}$, and returned if verified.
If no candidate clears the gate within $T$ iterations, the Pareto-optimal candidate with the highest aggregate rubric score is committed as a fallback.
The gate ensures that expensive Lean calls are reserved for decompositions that already satisfy the quality criteria predictive of downstream success.
Algorithm~\ref{alg:tomapo} summarizes the full procedure.

\begin{algorithm}[t]
\caption{\ToMap: Test-time Optimization for Multi-Agent Proof autoformalization}
\label{alg:tomapo}
\begin{algorithmic}[1]
\Require Input $x=(t,p)$, max iterations $T$, gate threshold $\epsilon$
\State Initialize pool $\mathcal{Z}_0 \leftarrow \{\D_{\theta_D}(x)\}$ via the initial Decomposer output
\For{$t = 1, \ldots, T$}
    \State Score each $z \in \mathcal{Z}_{t-1}$: compute $\rho(z)$ via LLM rubric judges
    \State Compute Pareto frontier $\mathcal{F}_{t-1}$
    \If{$\exists\, z \in \mathcal{F}_{t-1}$ s.t. $\min_k \rho_k(z) \geq \epsilon$}
        \State Commit $z$ to Formalizer-Prover; check with $\mathcal{V}$; \Return $\tau(x;\theta)$
    \EndIf
    \State Select parent $z^* \leftarrow \arg\max_{z \in \mathcal{F}_{t-1}} \sum_k \rho_k(z)$
    \State Derive critique $c \leftarrow \mathrm{Critique}(z^*, \rho(z^*), x)$
    \State Propose $\tilde{z} \leftarrow \mathrm{ReflectiveProposer}(z^*, c, x)$; compute $\rho(\tilde{z})$
    \State Update $\mathcal{Z}_t \leftarrow \mathrm{UpdatePool}(\mathcal{Z}_{t-1} \cup \{\tilde{z}\})$
\EndFor
\State Fallback: commit $z^* = \arg\max_{z \in \mathcal{F}_T} \sum_k \rho_k(z)$ to Formalizer-Prover
\State Check with $\mathcal{V}$; \Return $\tau(x;\theta)$
\end{algorithmic}
\end{algorithm}

\section{Experiments}
\label{sec:experiments}
We conduct a series of experiments and find that, compared with other full-proof autoformalization baselines, the translations generated by \ToMap achieve the highest semantic faithfulness, while also attaining the best overall syntactic–semantic performance. In Section~\ref{sec:Experimental_setup}, we introduce the experimental setup, where we align the inference budgets of substantially different reasoning pipelines as closely as possible and adopt a unified, reliable semantic evaluation protocol. In Section~\ref{sec:main_results}, we present the main experimental results. In Section~\ref{sec:ablation_study}, we conduct an ablation study on the number of optimization rounds to evaluate the efficiency of \ToMap.
\subsection{Experimental Setup}
\label{sec:Experimental_setup}
\paragraph{Benchmarks.}
We evaluate on two datasets comprising natural-language mathematical problems paired with their corresponding proofs: (1) miniF2F~\citep{miniF2F} , a widely used benchmark at the level of high-school mathematical competitions; we use its test split, which contains 244 problems. (2) \textsc{\textsc{ProofFlowBench}}~\citep{cabral2025proofflow}, an undergraduate-level mathematics dataset containing 184 problems. It is specifically designed to evaluate automated proof formalization pipelines, with an emphasis on proof-oriented problems rather than computational ones.
\paragraph{Baselines.}
We compare against existing methods for full-proof autoformalization: \textsc{ProofFlow}~\citep{cabral2025proofflow} performs step-wise proof formalization, where each stage retains the complete failure trace and uses it for subsequent retries. ProofBridge~\citep{jana2026proofbridge} trains both a retrieval model and a translation model, and employs a retrieval-augmented fine-tuned model to jointly translate, producing a single Lean proof. M2F~\citep{m2f}, a Codex-based verifier-certified refinement framework for project-level autoformalization. Monotonic~\citep{monotonic} directly optimizes the generated Lean code through iterative translation refinement, producing a single Lean proof. We also include two naive Codex-based baselines: Codex End2End and Codex D-Correction. The former prompts Codex to directly translate the complete natural-language proof into a single Lean proof, whereas the latter uses Codex as an agentic optimizer to iteratively self-correct the Decomposer for 10 iterations.

\paragraph{Evaluation metrics.}
We consider four evaluation metrics. (1) Syntactic Correctness (SC) measures whether the Lean code for a problem fully passes verification by the Lean compiler. (2) Semantic Faithfulness (SF) assesses whether the translation of the natural-language problem and its corresponding proof is semantically faithful. We largely follow the evaluation protocol of ReForm~\citep{reform}, which adopts an LLM-as-a-judge approach with Qwen3-235B-A22B~\citep{yang2025qwen3} as the judge model. The judgment is binary, reported as True or False, and this setup has been validated as a relatively reliable method for semantic evaluation. We modify the evaluation target from the natural-language problem alone to the natural-language problem together with its corresponding proof, and further require the Lean code to remain semantically faithful to each step of the proof. The evaluation criteria are provided in Appx~\ref{app:eval}. (3) Time denotes the wall-clock time spent on each problem, i.e., the total actual runtime of the task divided by the number of problems in the task. (4) Output Tokens (OT) denotes the total number of tokens generated by all LLM calls during the translation for each problem.
\paragraph{Model configuration.}
Detailed model configurations are provided in Appx~\ref{app:model_configuration}. \ToMap uses Goedel-Formalizer-V2-32B and Goedel-Prover-V2-32B as its formal language models~\citep{lin2025goedel}. We choose these two models to cover two representative model settings: Qwen3-30B-A3B-Instruct-2507~\citep{yang2025qwen3} serves as a strong open-weight model that can be locally deployed and reproduced, while Gemini3-Pro~\citep{team2023gemini} represents a proprietary frontier model accessed through APIs. \ToMap-Qwen30B and \ToMap-Gemini use Qwen3-30B-A3B-Instruct-2507 and Gemini3-Pro as the natural language model, respectively. The model configuration of \textsc{ProofFlow} is aligned with that of \ToMap-Gemini, while the other methods keep their original settings.
\paragraph{Inference budget and hyperparameter settings.}Due to substantial differences in the inference pipelines of different methods, it is difficult to compare all methods under a strictly identical inference budget. \textsc{ProofFlow} uses the same model configuration as our method. We measure the inference budget by the inference time on the same hardware. Since the prompt lengths of different workflows vary substantially, in parallel invocation scenarios, the time required to generate the same number of output tokens may differ significantly. We deploy each local model used in our experiments on a single NVIDIA A800 80GB GPU. Both methods perform parallel invocations, and we ensure that the degree of parallelism is sufficient to fully utilize the KV cache. For the remaining methods, we jointly control the running time and the number of output tokens to make the comparison as fair as possible. \ToMap{} is configured with 10 optimization iterations, and the gate threshold $\epsilon$ is set to 1. The detailed hyperparameters are provided in Appx~\ref{app:Hyperparameter_Settings}. The Formalizer and Prover in \textsc{ProofFlow} are each allowed up to 5 retries; ProofBridge reports Pass@64; and Monotonic performs 10 rounds of optimization.

\subsection{Main Results}
\label{sec:main_results}
\begin{table}[ht]
\centering
\caption{Main results on full-proof autoformalization benchmarks.}
\label{tab:main_results}

{\footnotesize
\setlength{\tabcolsep}{2.5pt}
\renewcommand{\arraystretch}{1.28}
\begin{tabular*}{\linewidth}{@{}l@{\extracolsep{\fill}}ccccc@{\hspace{8pt}}ccccc@{}}
\toprule
\multirow{2}{*}{\textbf{Method}}
& \multicolumn{5}{c}{\textbf{\textsc{ProofFlowBench}}}
& \multicolumn{5}{c}{\textbf{miniF2F}} \\
\cmidrule(lr){2-6}\cmidrule(lr){7-11}
& Time(s) & OT(k) & SC & SF & SC$\wedge$SF
& Time(s) & OT(k) & SC & SF & SC$\wedge$SF \\
\midrule

\rowcolor{gray!15}
\multicolumn{11}{c}{\textbf{Training-Based}} \\
\midrule
ProofBridge
& 113.48 & 87.39 & 92.93 & 9.24 & 9.24
& 61.23 & 63.80 & 95.08 & 29.51 & 28.69 \\

\midrule
\rowcolor{gray!15}
\multicolumn{11}{c}{\textbf{Training-Free}} \\
\midrule
Codex End2End
& 104.22 & 21.34 & 25.00 & 32.61 & {21.20}
& 83.62 & 28.92 & 49.18 & 58.20 & {47.54} \\

M2F
& 882.47 & 194.40 & 35.33 & 38.59 & 12.50
& 503.46 & 197.64 & 10.66 & 19.67 & 7.79 \\

Codex D-Correction
& 1554.50 & \maxcost{219.14} & 57.14 & 23.63 & 10.99
& \maxcost{1172.33} & \maxcost{199.84} & 71.72 & 33.61 & 18.44 \\

Monotonic
& 201.39 & 176.42 & \textbf{84.78} & 15.22 & 14.67
& 229.04 & 143.01 & \textbf{100.00} & 16.80 & 16.80 \\

\textsc{ProofFlow}
& \maxcost{2348.48} & 90.06 & 36.41 & 33.15 & 15.76
& 1125.73 & 49.28 & 61.89 & 46.31 & 35.25 \\

\ToMap-Qwen30B
& 610.10 & 98.81 & 66.30 & 27.71 & 21.10
& 513.44 & 83.33 & 86.89 & 34.02 & 29.92 \\

\ToMap-Gemini
& 850.27 & 139.25 & 63.59 & \best{51.09} & \best{40.22}\gain{19.0\%}
& 665.41 & 91.07 & 82.79 & \best{63.93} & \best{55.74}\gain{8.2\%} \\
\bottomrule
\end{tabular*}

\vspace{2pt}
\begin{minipage}{0.98\linewidth}
\scriptsize
\emph{Notes.}
SC denotes syntactic correctness, SF denotes semantic faithfulness, and
SC$\wedge$SF denotes the percentage of examples satisfying both criteria.
Time is the average wall-clock time per problem, and OT denotes output tokens in thousands.
Bold numbers mark the best semantic and joint syntactic-semantic scores.
$^{\dagger}$ marks the largest value in each cost column, i.e., Time(s) or OT(k), within the corresponding benchmark.
For training-based methods, reported time is inference-only and does not include training cost.
\textsuperscript{*}\ToMap{}-Qwen30B and \ToMap{}-Gemini use
Qwen3-30B-A3B-Instruct-2507 and Gemini3-Pro, respectively, as the natural-language model.
\end{minipage}

}

\end{table}

The results are presented in Table~\ref{tab:main_results}. Our method achieves the highest scores on Semantic Faithfulness and Syntactic Correctness $\wedge$ Semantic Faithfulness. Methods that generate a single monolithic Lean proof can achieve relatively high SC scores, but they struggle to faithfully translate the individual steps of natural-language proofs. In practice, these methods appear to use the natural-language proof primarily as guidance for producing Lean tactics that pass verification, rather than as a structure to be translated faithfully. Moreover, they may generate Lean statements that are semantically incorrect and simplified, which can unduly inflate their Syntactic Correctness scores. By contrast, step-wise proof formalization better preserves semantic faithfulness by explicitly aligning the formalization process with intermediate proof steps. \ToMap{} optimization substantially improves the quality of the decomposition; specific examples are provided in Appx~\ref{app:Decomposition_Optimization_Examples}. 
Compared the agentic baseline, Codex End2End, our method achieves absolute score gains of 38.59\% on \textsc{ProofFlowBench} and 33.61\% on miniF2F for SC, 18.48\% and 5.73\% for SF, and 19.02\% and 8.20\% for SC$\wedge$SF, respectively.

\subsection{Ablation Study}
\label{sec:ablation_study}
We instantiate the Decomposer in \ToMap with Qwen3-30B-A3B-Instruct-2507 and Gemini3-Pro as the natural-language model, respectively.
The stronger Gemini3-Pro setting yields substantially higher Semantic Faithfulness, while the two settings achieve comparable Syntactic Correctness.
This contrast suggests that stronger natural-language reasoning mainly improves whether the generated formal proof follows the original informal proof, whereas the remaining syntactic failures are increasingly tied to the fixed formal-language executor, including statement formalization and tactic generation.

We further study the effect of the maximum number of GEPA-style optimization rounds in Table~\ref{tab:ablation_rounds}.
Increasing the round budget consistently improves Syntactic Correctness, but also increases wall-clock cost.
This exposes a clear accuracy-cost trade-off: most gains are obtained within a small number of decomposition-evolution rounds, after which additional rounds provide diminishing returns.
The maximum round budget, therefore, serves as a practical test-time hyperparameter: smaller budgets are preferable when latency or cost is constrained, while larger budgets can be used when maximizing verification success is the primary objective.

\newcommand{\scorenogain}[1]{%
  \makebox[7.0em][c]{%
    \makebox[3.0em][r]{#1}%
    \hspace{1.5em}%
    \makebox[3.5em][l]{}%
  }%
}

\newcommand{\scorewithgain}[2]{%
  \makebox[7.0em][c]{%
    \makebox[3.0em][r]{#1}%
    \hspace{1.5em}%
    \makebox[3.5em][l]{\textcolor{red}{\gain{#2\%}}}%
  }%
}

\begin{table}[ht]
\centering
\caption{Syntactic correctness under different maximum rounds.}
\label{tab:ablation_rounds}
\setlength{\tabcolsep}{6pt}
\renewcommand{\arraystretch}{1.15}

\begin{tabular*}{\linewidth}{@{}l@{\extracolsep{\fill}}ccc@{}}
\toprule
\multirow{2}{*}{\textbf{Method}}
& \multirow{2}{*}{\textbf{Max Rounds}}
& \multicolumn{1}{c}{\textbf{\textsc{ProofFlowBench}}}
& \multicolumn{1}{c}{\textbf{miniF2F}} \\
\cmidrule(lr){3-3}\cmidrule(lr){4-4}
& & Syntactic Correctness & Syntactic Correctness \\
\midrule
\multirow{4}{*}{\ToMap-Qwen30B}
& 1  & \scorenogain{39.67} & \scorenogain{57.78} \\
& 3  & \scorewithgain{52.72}{13.05} & \scorewithgain{75.41}{17.63} \\
& 5  & \scorewithgain{59.24}{6.52}  & \scorewithgain{81.15}{5.74} \\
& 10 & \scorewithgain{66.30}{7.06}  & \scorewithgain{86.89}{5.74} \\
\midrule
\multirow{4}{*}{\ToMap-Gemini}
& 1  & \scorenogain{35.33} & \scorenogain{61.89} \\
& 3  & \scorewithgain{50.54}{15.21} & \scorewithgain{75.41}{13.52} \\
& 5  & \scorewithgain{56.52}{5.98}  & \scorewithgain{79.09}{3.68} \\
& 10 & \scorewithgain{63.59}{7.07}  & \scorewithgain{82.79}{3.70} \\
\bottomrule
\end{tabular*}

\end{table}

\section{Related work}

\paragraph{Autoformalization and full-proof formalization.} Autoformalization with Large Language Models~\citep{wu2022autoformalization} studies LLM-based translation from informal mathematics to formal languages such as Isabelle and Lean. Recent work moves from theorem-only translation toward full-proof autoformalization. \textsc{ProofFlow}~\citep{cabral2025proofflow} constructs dependency graphs for faithful proof autoformalization, ProofBridge~\citep{jana2026proofbridge} aligns natural-language and formal proof representations through joint embeddings, M2F~\citep{m2f} targets autoformalization of mathematical literature at scale, and Monotonic ~\citep{monotonic} studies iterative improvement for autoformalization without relying on formal references. \ToMap{} performs step-wise proof autoformalization and iteratively optimizes the pipeline by targeting the Decomposer.

\paragraph{Weak-link and causal analysis of multi-agent systems.} Agent systems study how autonomous agents perceive task contexts, interact with environments, and act to solve problems, with multi-agent compositions further introducing challenges in coordination, communication, and role specialization~\citep{liu2025semantic,zhang2024whale,yuan2023survey}. Recent LLM multi-agent systems study scalable cooperation through role-playing communicative agents and conversational multi-agent orchestration~\citep{li2023camel,wu2023autogen,hong2024metagpt}.
In formal mathematics, agent-based designs further modularize theorem proving into staged interactions with proof assistants~\citep{baba2025prover}.
Complementary to these system templates, recent work explicitly targets bottleneck agents and analyzes component effects via interventions in collaborative LLM systems~\citep{weak-link,causal_analysis}.
Building on this diagnostic perspective, we compare matched-interface edits to the Decomposer, Formalizer, and Prover under identical Lean verifier feedback to identify the most feedback-recoverable stage before allocating test-time optimization.

\paragraph{GEPA-style prompt optimization.} Automatic prompt improvement spans candidate search and selection~\citep{zhou2022large}, iterative optimization of natural-language instructions~\citep{yang2024large}, evolutionary prompt updates~\citep{fernando2024promptbreeder}, and trajectory-level reflective revision~\citep{shinn2023reflexion}.
GEPA~\citep{2025gepa} further scales reflective evolution to compound systems using trajectory-level textual feedback and Pareto maintenance over candidates.
Closer to autoformalization, ReForm~\citep{reform} studies reflective refinement under bounded prospective optimization.
\ToMap instantiates this family for full-proof pipelines by evolving Decomposer-side prompts and proof-unit specifications under dense rubric signals, and only commits promising candidates to frozen Formalizer-Prover execution with Lean verification.

\section{Conclusion and Limitation}
Full-proof autoformalization connects natural-language mathematical proofs with machine-checkable formal reasoning. 
We study it as a multi-agent LLM system grounded by Lean verification, where decomposition, formalization, and proof search are handled by specialized agents. 
Our weak-link analysis identifies the Decomposer as the most effective interface to optimize under a fixed Formalizer-Prover executor, since faithful, self-contained, and dependency-aware proof units greatly simplify downstream proof construction.
We introduce \ToMap{}, a GEPA-style test-time optimizer that evolves proof decompositions. 
This focuses limited test-time compute on improving the decomposition rather than repeatedly retrying the full pipeline. 
Experiments show that \ToMap{} improves end-to-end Lean verification and semantic faithfulness on full-proof autoformalization benchmarks.

Limitations remain. 
Our evaluation is restricted to benchmark-scale proofs rather than long, research-level mathematical proofs. 
Moreover, we assume that the input informal proofs are correct; extending the system to detect and repair incorrect or incomplete proof steps is left for future work.

\clearpage

\bibliography{references}


\newpage
\appendix
\onecolumn

 \section{Prompt}
\label{app:prompt}
This section presents the prompts used by the agents in \ToMap{}, including the decomposer for natural-language problem decomposition~\ref{app:decomposer}, the formalizer for translating natural-language subclaims into Lean statements~\ref{app:formalizer}, the prover for generating Lean proofs that complete Lean statements~\ref{app:prover}, the GEPA judge for evaluating natural-language decompositions~\ref{app:gepa_judge}, and the GEPA rewriter for revising natural-language decompositions~\ref{app:gepa_rewriter}. It also describes the semantic evaluation criteria used in the evaluation process~\ref{app:eval}.

\tcbset{
    decomposerstyle/.style={
        enhanced,
        breakable,
        colback=blue!5,
        colframe=blue!60!black!70,
        boxrule=0.8pt,
        arc=4mm,
        left=4mm,
        right=4mm,
        top=2mm,
        bottom=2mm,
        fonttitle=\bfseries,
        title=Decomposer,
        before skip=6pt,
        after skip=6pt
    },
    formalizerstyle/.style={
        enhanced,
        breakable,
        colback=blue!5,
        colframe=blue!60!black!70,
        boxrule=0.8pt,
        arc=4mm,
        left=4mm,
        right=4mm,
        top=2mm,
        bottom=2mm,
        fonttitle=\bfseries,
        title=Formalizer,
        before skip=6pt,
        after skip=6pt
    },
    proverstyle/.style={
        enhanced,
        breakable,
        colback=blue!5,
        colframe=blue!60!black!70,
        boxrule=0.8pt,
        arc=4mm,
        left=4mm,
        right=4mm,
        top=2mm,
        bottom=2mm,
        fonttitle=\bfseries,
        title=Prover,
        before skip=6pt,
        after skip=6pt
    },
    gepajudgestyle/.style={
        enhanced,
        breakable,
        colback=blue!5,
        colframe=blue!60!black!70,
        boxrule=0.8pt,
        arc=4mm,
        left=4mm,
        right=4mm,
        top=2mm,
        bottom=2mm,
        fonttitle=\bfseries,
        title=GEPA Judge,
        before skip=6pt,
        after skip=6pt
    },
    geparewriterstyle/.style={
        enhanced,
        breakable,
        colback=blue!5,
        colframe=blue!60!black!70,
        boxrule=0.8pt,
        arc=4mm,
        left=4mm,
        right=4mm,
        top=2mm,
        bottom=2mm,
        fonttitle=\bfseries,
        title=GEPA Rewriter,
        before skip=6pt,
        after skip=6pt
    },
    evalstyle/.style={
        enhanced,
        breakable,
        colback=blue!5,
        colframe=blue!60!black!70,
        boxrule=0.8pt,
        arc=4mm,
        left=4mm,
        right=4mm,
        top=2mm,
        bottom=2mm,
        fonttitle=\bfseries,
        title=Evaluation Criteria,
        before skip=6pt,
        after skip=6pt
    }
}
\subsection{Decomposer}
\label{app:decomposer}
\begin{tcolorbox}[decomposerstyle]
\begingroup
\raggedright
\sloppy
\noindent{}You are an expert at analyzing mathematical proofs and creating structured proof chains for Lean4 formalization. Given a natural language theorem and proof, your task is to generate a structured proof chain in JSON format that exactly represents the logical structure of the proof.\\
\par\smallskip
\noindent{}---\\
\par\smallskip
\noindent{}\#\#\# Key Requirements for Lean-Friendly Formalization\\
\par\smallskip
\noindent{}1. **Contextual Text Extraction**\\
\par\smallskip
\noindent{}\hspace*{1.5em}* The \textasciigrave{}"natural\_language"\textasciigrave{} field must contain an **exact quote** from the proof text justifying the lemma's inference.\\
\noindent{}\hspace*{1.5em}* The \textasciigrave{}"statement"\textasciigrave{} field should be **self-contained** and **explicit** \textemdash{} repeat necessary assumptions from earlier steps, and reference them clearly using their IDs.\\
\par\smallskip
\noindent{}2. **Step-by-Step Granularity (CRITICAL: Maximize Prover Success Rate)**\\
\par\smallskip
\noindent{}\hspace*{1.5em}* **Decompose the proof into VERY small, atomic steps**. Each logical inference must be a **separate lemma**.\\
\noindent{}\hspace*{1.5em}* **PRIORITY: Break down complex steps into simpler sub-steps** that can each be proven with basic tactics (simp, ring, linarith, exact).\\
\noindent{}\hspace*{1.5em}* If a step requires multiple reasoning steps, **split it into multiple lemmas** even if they seem trivial. It's better to have more, simpler lemmas than fewer, complex ones.\\
\noindent{}\hspace*{1.5em}* **When in doubt, decompose further**: A lemma that might be provable with \textasciigrave{}simp\textasciigrave{} or \textasciigrave{}ring\textasciigrave{} alone is much more likely to succeed than one requiring complex reasoning.\\
\noindent{}\hspace*{1.5em}* Only merge steps when they can be computed with a **single simple tactic** (like \textasciigrave{}simp\textasciigrave{}, \textasciigrave{}ring\textasciigrave{}, or \textasciigrave{}linarith\textasciigrave{}).\\
\par\smallskip
\noindent{}3. **Formalization Using Lean's Tactics (PRIORITIZE SIMPLE TACTICS)**\\
\par\smallskip
\noindent{}\hspace*{1.5em}* For each lemma, you must include a **Lean hint** (\textasciigrave{}lean\_hint\textasciigrave{}), suggesting appropriate tactics or actions.\\
\noindent{}\hspace*{1.5em}* **TACTIC PRIORITY ORDER (use the simplest possible tactic)**:\\
\noindent{}\hspace*{2.5em}\\
\noindent{}\hspace*{2.5em}**TIER 1 (PREFERRED - Highest success rate):**\\
\noindent{}\hspace*{2.5em}* \textasciigrave{}simp\textasciigrave{} - For basic simplifications\\
\noindent{}\hspace*{2.5em}* \textasciigrave{}ring\textasciigrave{} - For ring/field arithmetic\\
\noindent{}\hspace*{2.5em}* \textasciigrave{}linarith\textasciigrave{} - For linear arithmetic\\
\noindent{}\hspace*{2.5em}* \textasciigrave{}exact\textasciigrave{} - When the goal matches an assumption exactly\\
\noindent{}\hspace*{2.5em}* \textasciigrave{}norm\_num\textasciigrave{} - For numerical computations\\
\noindent{}\hspace*{2.5em}\\
\noindent{}\hspace*{2.5em}**TIER 2 (Use when Tier 1 is insufficient):**\\
\noindent{}\hspace*{2.5em}* \textasciigrave{}rw\textasciigrave{} - For rewriting with equalities\\
\noindent{}\hspace*{2.5em}* \textasciigrave{}apply\textasciigrave{} - For applying theorems\\
\noindent{}\hspace*{2.5em}* \textasciigrave{}use\textasciigrave{} - For existential goals\\
\noindent{}\hspace*{2.5em}\\
\noindent{}\hspace*{2.5em}**TIER 3 (Avoid if possible - Lower success rate):**\\
\noindent{}\hspace*{2.5em}* \textasciigrave{}by\_cases\textasciigrave{} - Case analysis (split into separate lemmas instead)\\
\noindent{}\hspace*{2.5em}* \textasciigrave{}cases\textasciigrave{} - Pattern matching (split into separate lemmas instead)\\
\noindent{}\hspace*{2.5em}* \textasciigrave{}by\_contra\textasciigrave{} - Proof by contradiction (split into separate lemmas instead)\\
\noindent{}\hspace*{2.5em}* \textasciigrave{}induction\textasciigrave{} - Induction (split into separate lemmas instead)\\
\noindent{}\hspace*{2.5em}\\
\noindent{}\hspace*{1.5em}* **CRITICAL RULE**: If a lemma requires tactics from Tier 2 or 3, consider splitting it into multiple simpler lemmas that can each use Tier 1 tactics.\\
\noindent{}\hspace*{1.5em}* **For algebraic manipulations**: Prefer \textasciigrave{}ring\textasciigrave{} or \textasciigrave{}simp\textasciigrave{} over \textasciigrave{}ring\_nf\textasciigrave{} or complex \textasciigrave{}rw\textasciigrave{} chains.\\
\noindent{}\hspace*{1.5em}* **For direct applications**: Prefer \textasciigrave{}exact\textasciigrave{} over \textasciigrave{}apply\textasciigrave{} when possible.\\
\par\smallskip
\noindent{}4. **Managing Dependencies**\\
\noindent{}\hspace*{1.5em}The "statement" field restates all required assumptions explicitly, so each lemma is self-contained and can be understood by Lean without relying on hidden context.\\
\par\smallskip
\noindent{}5. **Lean-Friendly Statement Format**\\
\par\smallskip
\noindent{}\hspace*{1.5em}* For **each lemma**:\\
\par\smallskip
\noindent{}\hspace*{2.5em}* **Start with assumptions**: Explicitly restate all prior lemmas or assumptions.\\
\noindent{}\hspace*{2.5em}* **State the conclusion**: Clearly state what the lemma establishes.\\
\noindent{}\hspace*{1.5em}* **Example format**:\\
\par\smallskip
\noindent{}\hspace*{2.5em}\textasciigrave{}\textasciigrave{}\textasciigrave{}json\\
\noindent{}\hspace*{2.5em}\{\\
\noindent{}\hspace*{3.5em}"id": "lk",\\
\noindent{}\hspace*{3.5em}"natural\_language": "[exact text from proof]",\\
\noindent{}\hspace*{3.5em}"statement": "We assume:\textbackslash{}\textbackslash{}n\textbullet{} [all dependencies with IDs]\textbackslash{}\textbackslash{}nTherefore, we conclude:\textbackslash{}\textbackslash{}n\textbullet{} [the new fact] [lk]",\\
\noindent{}\hspace*{3.5em}"dependencies": ["tc\_1", "def\_1"],\\
\noindent{}\hspace*{3.5em}"lean\_hint": "[brief tactic]"\\
\noindent{}\hspace*{2.5em}\}\\
\noindent{}\hspace*{2.5em}\textasciigrave{}\textasciigrave{}\textasciigrave{}\\
\noindent{}\hspace*{1.5em}* **For definitions** (\textasciigrave{}def\_k\textasciigrave{}), provide a simple statement format:\\
\par\smallskip
\noindent{}\hspace*{2.5em}\textasciigrave{}\textasciigrave{}\textasciigrave{}json\\
\noindent{}\hspace*{2.5em}\{\\
\noindent{}\hspace*{3.5em}"id": "def\_k",\\
\noindent{}\hspace*{3.5em}"natural\_language": "[exact definition from proof]",\\
\noindent{}\hspace*{3.5em}"statement": "Definition:\textbackslash{}\textbackslash{}n\textbullet{} [definition content] [def\_k]",\\
\noindent{}\hspace*{3.5em}"dependencies": ["tc\_1"]\\
\noindent{}\hspace*{2.5em}\}\\
\noindent{}\hspace*{2.5em}\textasciigrave{}\textasciigrave{}\textasciigrave{}\\
\par\smallskip
\noindent{}6. **No Error Correction**\\
\par\smallskip
\noindent{}\hspace*{1.5em}* **Do not introduce new logical steps** that are not explicitly present in the natural language proof. If a proof is incomplete or incorrect, **represent it exactly as it is written**.\\
\par\smallskip
\par\smallskip
\par\smallskip
\noindent{}\#\#\# Node Types\\
\par\smallskip
\noindent{}There are **four node types** in the structured proof chain:\\
\par\smallskip
\noindent{}1. **Theorem Condition (\textasciigrave{}tc\_k\textasciigrave{})** \textemdash{} Assumptions and given conditions, **only formalized** (no proving).\\
\noindent{}2. **Theorem Solution (\textasciigrave{}ts\_k\textasciigrave{})** \textemdash{} The final conclusion of the proof, **requires formalization and proof**.\\
\noindent{}3. **Definition (\textasciigrave{}def\_k\textasciigrave{})** \textemdash{} Definitions, assumptions, and auxiliary results that are not part of the proof but needed for formalization.\\
\noindent{}4. **Lemma (\textasciigrave{}lk\textasciigrave{})** \textemdash{} Intermediate proof steps that require formalization and Lean tactics.\\
\par\smallskip
\noindent{}\#\#\# Formalization Guidelines for Lemmas (Maximize Prover Success)\\
\par\smallskip
\noindent{}**What to Capture:**\\
\par\smallskip
\noindent{}* Each **atomic logical inference** must be captured as a **separate lemma**.\\
\noindent{}* **Decompose aggressively**: If a step involves multiple operations (e.g., substitution + simplification + equality), split it into separate lemmas for each operation.\\
\noindent{}* For each lemma:\\
\par\smallskip
\noindent{}\hspace*{1.0em}* Restate **all assumptions** (dependencies) explicitly, even if they are already known.\\
\noindent{}\hspace*{1.0em}* Express **the conclusion** of the lemma in mathematical terms.\\
\noindent{}\hspace*{1.0em}* **Lean Hint**: **ALWAYS prefer the simplest possible tactic** from Tier 1 (simp, ring, linarith, exact, norm\_num).\\
\noindent{}\hspace*{1.0em}* **If you cannot suggest a Tier 1 tactic, the lemma is likely too complex** - consider splitting it further.\\
\noindent{}\hspace*{1.0em}* **Provide detailed hints**: The \textasciigrave{}lean\_hint\textasciigrave{} should include:\\
\noindent{}\hspace*{2.0em}- The specific tactic to use (e.g., \textasciigrave{}ring\textasciigrave{}, \textasciigrave{}simp\textasciigrave{}, \textasciigrave{}linarith\textasciigrave{})\\
\noindent{}\hspace*{2.0em}- Key intermediate steps if needed (e.g., "First use \textasciigrave{}rw\textasciigrave{} to substitute, then \textasciigrave{}simp\textasciigrave{}")\\
\noindent{}\hspace*{2.0em}- Important lemmas or facts to reference (e.g., "Use \textasciigrave{}add\_comm\textasciigrave{} and \textasciigrave{}add\_assoc\textasciigrave{}")\\
\noindent{}\hspace*{2.0em}- Any algebraic identities or simplifications needed\\
\noindent{}\hspace*{2.0em}- **For complex lemmas**: Break down the proof strategy into clear steps (e.g., "1. Use \textasciigrave{}rw\textasciigrave{} with dependency [l5], 2. Apply \textasciigrave{}ring\textasciigrave{} to simplify, 3. Use \textasciigrave{}linarith\textasciigrave{} to conclude")\\
\par\smallskip
\noindent{}**Examples of Good Decomposition:**\\
\par\smallskip
\noindent{}*  **Bad**: One lemma that says "Substitute x = a+b into f(x) and simplify to get f(a+b) = g(a,b)"\\
\noindent{}\hspace*{1.0em}*  **Good**: \\
\noindent{}\hspace*{2.0em}- Lemma 1: "If x = a+b, then f(x) = f(a+b)" (hint: \textasciigrave{}rw [h\_x\_eq]\textasciigrave{} where \textasciigrave{}h\_x\_eq\textasciigrave{} is the assumption)\\
\noindent{}\hspace*{2.0em}- Lemma 2: "f(a+b) = g(a,b)" (hint: \textasciigrave{}simp [f\_def]\textasciigrave{} or \textasciigrave{}ring\textasciigrave{} to simplify the expression)\\
\par\smallskip
\noindent{}*  **Bad**: One lemma that says "By case analysis, either P or not P, and in both cases we get Q"\\
\noindent{}\hspace*{1.0em}*  **Good**:\\
\noindent{}\hspace*{2.0em}- Lemma 1: "If P, then Q" (hint: \textasciigrave{}exact h\_P\_implies\_Q\textasciigrave{} or \textasciigrave{}simp [h\_P]\textasciigrave{})\\
\noindent{}\hspace*{2.0em}- Lemma 2: "If not P, then Q" (hint: \textasciigrave{}exact h\_notP\_implies\_Q\textasciigrave{} or \textasciigrave{}simp [h\_notP]\textasciigrave{})\\
\noindent{}\hspace*{2.0em}- Lemma 3: "Q" (hint: \textasciigrave{}by\_cases h : P; [exact Lemma1 h, exact Lemma2 h]\textasciigrave{})\\
\par\smallskip
\noindent{}*  **Bad**: One lemma that says "From W + Y = X + Z, we get X - W = Z - Y"\\
\noindent{}\hspace*{1.0em}*  **Good**:\\
\noindent{}\hspace*{2.0em}- Lemma 1: "W + Y = X + Z implies W + Y - X - Z = 0" (hint: \textasciigrave{}linarith [h\_eq]\textasciigrave{} where \textasciigrave{}h\_eq : W + Y = X + Z\textasciigrave{})\\
\noindent{}\hspace*{2.0em}- Lemma 2: "W + Y - X - Z = 0 implies (W - X) + (Y - Z) = 0" (hint: \textasciigrave{}ring\textasciigrave{} to rearrange)\\
\noindent{}\hspace*{2.0em}- Lemma 3: "(W - X) + (Y - Z) = 0 implies W - X = -(Y - Z)" (hint: \textasciigrave{}linarith\textasciigrave{})\\
\noindent{}\hspace*{2.0em}- Lemma 4: "W - X = -(Y - Z) implies X - W = Y - Z" (hint: \textasciigrave{}linarith\textasciigrave{} to negate both sides)\\
\par\smallskip
\noindent{}**What **Not** to Capture:**\\
\par\smallskip
\noindent{}* **Introductions**: Do not include introductions like "Let ( x \textbackslash{}in A )" or "Fix ( \textbackslash{}epsilon > 0 )".\\
\noindent{}* **Meta-comments**: Avoid phrases like "It suffices to show..." or "We proceed by induction".\\
\par\smallskip
\noindent{}---\\
\par\smallskip
\noindent{}\#\#\# Output Format\\
\par\smallskip
\noindent{}The output should be in **JSON format** and should follow the exact structure below. Ensure to follow this structure **without any unrelated content**.\\
\noindent{}Note that backslashes (\textbackslash{}) must be written as double backslashes (\textbackslash{}\textbackslash{}) in JSON strings.\\
\par\smallskip
\noindent{}\textasciigrave{}\textasciigrave{}\textasciigrave{}json\\
\noindent{}[\\
\noindent{}\hspace*{1.0em}\{\\
\noindent{}\hspace*{2.0em}"id": "tc\_1",\\
\noindent{}\hspace*{2.0em}"natural\_language": "[exact text from theorem statement]",\\
\noindent{}\hspace*{2.0em}"statement": "Premise:\textbackslash{}\textbackslash{}n\textbullet{} [mathematical content] [tc\_1]",\\
\noindent{}\hspace*{2.0em}"dependencies": []\\
\noindent{}\hspace*{1.0em}\},\\
\noindent{}\hspace*{1.0em}\{\\
\noindent{}\hspace*{2.0em}"id": "def\_1",\\
\noindent{}\hspace*{2.0em}"natural\_language": "[exact definition from proof]",\\
\noindent{}\hspace*{2.0em}"statement": "Definition:\textbackslash{}\textbackslash{}n\textbullet{} [definition content] [def\_1]",\\
\noindent{}\hspace*{2.0em}"dependencies": ["tc\_1"]\\
\noindent{}\hspace*{1.0em}\},\\
\noindent{}\hspace*{1.0em}\{\\
\noindent{}\hspace*{2.0em}"id": "l1",\\
\noindent{}\hspace*{2.0em}"natural\_language": "[exact text from proof]",\\
\noindent{}\hspace*{2.0em}"statement": "We assume:\textbackslash{}\textbackslash{}n\textbullet{} [restated content with IDs]\textbackslash{}\textbackslash{}nTherefore, we conclude:\textbackslash{}\textbackslash{}n\textbullet{} [new fact] [l1]",\\
\noindent{}\hspace*{2.0em}"dependencies": ["tc\_1", "def\_1"],\\
\noindent{}\hspace*{2.0em}"lean\_hint": "[brief tactic plan]"\\
\noindent{}\hspace*{1.0em}\},\\
\noindent{}\hspace*{1.0em}\{\\
\noindent{}\hspace*{2.0em}"id": "ts\_1",\\
\noindent{}\hspace*{2.0em}"natural\_language": "[final text]",\\
\noindent{}\hspace*{2.0em}"statement": "We assume:\textbackslash{}\textbackslash{}n\textbullet{} [all dependencies with IDs]\textbackslash{}\textbackslash{}nTherefore, we conclude:\textbackslash{}\textbackslash{}n\textbullet{} [theorem solution] [ts\_1]",\\
\noindent{}\hspace*{2.0em}"dependencies": ["tc\_1","l1", "def\_1"],\\
\noindent{}\hspace*{2.0em}"lean\_hint": "[brief tactic plan]"\\
\noindent{}\hspace*{1.0em}\}\\
\noindent{}]\\
\noindent{}\textasciigrave{}\textasciigrave{}\textasciigrave{}\\
\par\smallskip
\noindent{}\#\#\# The \textasciigrave{}statement\textasciigrave{} field (mandatory style)\\
\par\smallskip
\noindent{}* The **exact format** for lemmas and theorem solutions:\\
\par\smallskip
\noindent{}\hspace*{1.0em}\textasciigrave{}\textasciigrave{}\textasciigrave{}text\\
\noindent{}\hspace*{1.0em}We assume:\\
\noindent{}\hspace*{1.0em}\textbullet{} [actual content of tc\_1] [tc\_1];\\
\noindent{}\hspace*{1.0em}\textbullet{} [actual content of tc\_2] [tc\_2];\\
\noindent{}\hspace*{1.0em}\textbullet{} [actual content of l1] [l1];\\
\noindent{}\hspace*{1.0em}Therefore, we conclude:\\
\noindent{}\hspace*{1.0em}\textbullet{} [actual content of l2] [l2].\\
\noindent{}\hspace*{1.0em}\textasciigrave{}\textasciigrave{}\textasciigrave{}\\
\par\smallskip
\noindent{}* For **theorem conditions**:\\
\par\smallskip
\noindent{}\hspace*{1.0em}\textasciigrave{}\textasciigrave{}\textasciigrave{}text\\
\noindent{}\hspace*{1.0em}Premise:\\
\noindent{}\hspace*{1.0em}\textbullet{} [actual content of tc\_1] [tc\_1];\\
\noindent{}\hspace*{1.0em}\textasciigrave{}\textasciigrave{}\textasciigrave{}\\
\par\smallskip
\noindent{}* For **definitions**:\\
\par\smallskip
\noindent{}\hspace*{1.0em}\textasciigrave{}\textasciigrave{}\textasciigrave{}text\\
\noindent{}\hspace*{1.0em}Definition:\\
\noindent{}\hspace*{1.0em}\textbullet{} [actual content of def\_1] [def\_1];\\
\noindent{}\hspace*{1.0em}\textasciigrave{}\textasciigrave{}\textasciigrave{}\\
\par\smallskip
\noindent{}---\\
\par\smallskip
\noindent{}\#\#\# Quick Checklist (Prioritize Prover Success)\\
\par\smallskip
\noindent{}* **Be clear and complete**: Each step should include all necessary information (variables, assumptions, etc.), so Lean can follow it easily.\\
\noindent{}* **Keep it VERY simple**: Break the proof into **as many small steps as possible**. Prefer many simple lemmas over fewer complex ones.\\
\noindent{}* **Maximize Tier 1 tactics**: Each lemma should ideally be provable with \textasciigrave{}simp\textasciigrave{}, \textasciigrave{}ring\textasciigrave{}, \textasciigrave{}linarith\textasciigrave{}, \textasciigrave{}exact\textasciigrave{}, or \textasciigrave{}norm\_num\textasciigrave{}.\\
\noindent{}* **Decompose complex steps**: If a step involves substitution, simplification, and conclusion, split it into separate lemmas.\\
\noindent{}* **Avoid complex reasoning**: If a lemma requires case analysis, contradiction, or induction, consider splitting it into multiple simpler lemmas.\\
\noindent{}* **Stay true to the original**: Follow the author's reasoning exactly\textemdash{}don't change the logic, and don't add new concepts or ideas.\\
\noindent{}* **Reflect the proof as written**: Even if the proof has gaps or mistakes, represent it faithfully in Lean.\\
\noindent{}* **Test your hints**: If you suggest a tactic from Tier 2 or 3, ask yourself: "Can I split this into simpler lemmas that use Tier 1 tactics?"\\
\#\# Your Task

Now, please generate the proof decomposition for the following theorem and proof:

**Theorem:** \{ nl\_theorem \}

**Proof:** \{ nl\_proof \}

\endgroup
\end{tcolorbox}
\subsection{Formalizer}
\label{app:formalizer}
\begin{tcolorbox}[formalizerstyle]
\begingroup
\raggedright
\sloppy

\noindent{}You are a **thinking model** specialized in turning natural-language math statements into **Lean 4** code.\par
\smallskip\par
\noindent{}You will receive:\par
\smallskip\par
\noindent{}1. A lemma name (\textasciigrave{}lemma\_header\textasciigrave{}, e.g. \textasciigrave{}lemma l3\textasciigrave{}),\par
\noindent{}2. A natural-language statement for this proof step,\par
\noindent{}3. Name of dependencies\par
\noindent{}4. A Lean code skeleton with placeholders\par
\noindent{}5. **full Lean 4 code of prior steps** (``dependencies'')\par
\smallskip\par
\noindent{}Your job is to **emit exactly one Lean 4 code block** that compiles after replacing placeholders, while **not reproducing any prior lemmas/defs**. Treat prior steps only as hypotheses (see ``Using Dependencies'').\par
\smallskip\par
\noindent{}---\par
\smallskip\par
\noindent{}\#\# Hard Rules (highest priority)\par
\smallskip\par
\noindent{}**0. Single lemma only (no extras)**\par
\smallskip\par
\noindent{}* Output must contain **exactly one lemma/theorem definition** (the \textasciigrave{}lemma\_header\textasciigrave{}).\par
\noindent{}* The fenced block must contain **exactly one** of: \textasciigrave{}lemma \textasciigrave{} *or* \textasciigrave{}theorem \textasciigrave{}.\par
\noindent{}* The block must contain **exactly one** occurrence of \textasciigrave{}:= by\textasciigrave{} and of \textasciigrave{}sorry\textasciigrave{}.\par
\smallskip\par
\noindent{}1. **Provide one fenced Lean block**\par
\noindent{}\hspace*{1.0em}* /think /think Please think carefully before giving the final answer.\par
\noindent{}\hspace*{1.0em}* At the end, write **exactly one** fenced block: start with \textasciigrave{}lean4 and end with \textasciigrave{}.\par
\smallskip\par
\noindent{}2. **Header is fixed** (keep exactly this header in this order):\par
\smallskip\par
\noindent{}\textasciigrave{}\textasciigrave{}\textasciigrave{}lean4\par
\noindent{}import Mathlib\par
\noindent{}import Aesop\par
\smallskip\par
\noindent{}set\_option maxHeartbeats 0\par
\smallskip\par
\noindent{}open BigOperators Real Nat Topology Rat Filter\par
\noindent{}\textasciigrave{}\textasciigrave{}\textasciigrave{}\par
\smallskip\par
\noindent{}3. **Proof body format**\par
\smallskip\par
\noindent{}\hspace*{1.5em}* The proof body must be **exactly** \textasciigrave{}:= by\textbackslash{}nsorry\textasciigrave{}.\par
\smallskip\par
\noindent{}4. **Lemma name unchanged**\par
\smallskip\par
\noindent{}\hspace*{1.5em}* Keep the \textasciigrave{}lemma\_header\textasciigrave{} name and structure exactly; only fill in parameters/hypotheses and the goal.\par
\smallskip\par
\noindent{}5. **Prohibited content inside the block**: additional \textasciigrave{}lemma\textasciigrave{}, \textasciigrave{}theorem\textasciigrave{}, \textasciigrave{}def\textasciigrave{}, \textasciigrave{}example\textasciigrave{}, \textasciigrave{}instance\textasciigrave{}, \textasciigrave{}axiom\textasciigrave{}, \textasciigrave{}structure\textasciigrave{}, or extra \textasciigrave{}import\textasciigrave{}/\textasciigrave{}set\_option\textasciigrave{}/\textasciigrave{}open\textasciigrave{} sections beyond the single required header.\par
\smallskip\par
\noindent{}---\par
\smallskip\par
\noindent{}\#\# Using Dependencies\par
\smallskip\par
\noindent{}Often you are given the **full Lean code** of earlier steps (``dependencies'').\par
\noindent{}These prior lemmas contain **relevant Lean 4 information** that you may use when formalizing the current step, including:\par
\smallskip\par
\noindent{}* **Type declarations** (e.g., \textasciigrave{}\{n : \ensuremath{\mathbb{N}}\}\textasciigrave{}, \textasciigrave{}(A B : Set \ensuremath{\alpha})\textasciigrave{}, \textasciigrave{}(x : \ensuremath{\alpha})\textasciigrave{}, etc.)\par
\noindent{}* **Formalized statements (goals) of the previous lemmas** \textemdash{} i.e., their proposition types\par
\noindent{}* Any **constants and structures** appearing in those lemmas\par
\smallskip\par
\noindent{}**What you must do with dependencies:**\par
\smallskip\par
\noindent{}1. **Include every dependency as a named hypothesis** in the current lemma.\par
\noindent{}\hspace*{1.5em}* Use the **dependency lemma's name** as the hypothesis name.\par
\noindent{}\hspace*{1.5em}* Its hypothesis type is the **proposition type of that prior lemma**, adapted to match the variables in the current lemma.\par
\noindent{}\hspace*{1.5em}* Include all dependencies as hypotheses even if you think they are not relevant or will be used.\par
\smallskip\par
\noindent{}2. **Not mandatory to reproduce dependency code verbatim.** You are given prior code for *reference only*. \par
\noindent{}\hspace*{1.5em}* If needed, **adapt** the dependency statements to properly fit the current lemma's parameters and context.\par
\noindent{}\hspace*{1.5em}* Instantiate with the current variables when possible, or generalize with \ensuremath{\forall} if necessary.\par
\noindent{}\hspace*{1.5em}* Minor modifications (e.g., renaming variables, adapting types) are OK if needed for consistency.\par
\smallskip\par
\noindent{}3. **Preferred**: If the current lemma declares matching variables (same names/types), **instantiate** the dependency with those variables (no \ensuremath{\forall}).\par
\noindent{}\hspace*{2.5em}* Example: prior \textasciigrave{}lemma l1 (a b : \ensuremath{\mathbb{R}}) : a + b = b + a\textasciigrave{}, then you add \textasciigrave{}(l1 : a + b = b + a)\textasciigrave{} after declaring \textasciigrave{}(a b : \ensuremath{\mathbb{R}})\textasciigrave{}.\par
\smallskip\par
\noindent{}4. **Extra hypotheses are allowed**\par
\noindent{}You may add new hypothesis, such astype/variable declarations (e.g., \{\ensuremath{\alpha} : Type*\}, (n : \ensuremath{\mathbb{N}}), (x : \ensuremath{\mathbb{R}})) if they are needed to make the current lemma well-formed. These are in addition to the dependencies. Always place them before dependency hypotheses in the parameter list.\par
\smallskip\par
\noindent{}5. **Do not duplicate** a dependency that you've already added.\par
\smallskip\par
\noindent{}6. If no dependencies are provided, skip this section and proceed normally.\par
\smallskip\par
\noindent{}---\par
\smallskip\par
\noindent{}\#\# Parameter/Hypothesis Ordering (skeleton has none \ensuremath{\to} write from scratch)\par
\smallskip\par
\noindent{}When you fill the skeleton parameters, **place items in this order**:\par
\smallskip\par
\noindent{}1. **Type/implicit parameters first** (e.g., \textasciigrave{}\{n : \ensuremath{\mathbb{N}}\}\textasciigrave{}, \textasciigrave{}\{\ensuremath{\alpha} : Type*\}\textasciigrave{})\par
\noindent{}2. **Explicit variable declarations next** (e.g., \textasciigrave{}(A B : Set \ensuremath{\alpha}) (x : \ensuremath{\alpha}) (a b : \ensuremath{\mathbb{R}})\textasciigrave{})\par
\noindent{}3. **Dependency hypotheses** (one per prior lemma, named by lemma name; adapted as needed)\par
\smallskip\par
\noindent{}Then place a colon \textasciigrave{}:\textasciigrave{} and the **formalized goal statement**, followed by \textasciigrave{}:= by\textbackslash{}nsorry\textasciigrave{}.\par
\smallskip\par
\noindent{}---\par
\smallskip\par
\noindent{}\#\# Goal Formalization\par
\smallskip\par
\noindent{}* Replace \textasciigrave{}[place correct hypothesis here]\textasciigrave{} and \textasciigrave{}[place goal here]\textasciigrave{} with the **precise Lean proposition** matching the natural-language statement.\par
\noindent{}* The goal is to replicate the logic in the natural language statement in Lean4, so please preserve the meaning of the natural language proof.\par
\noindent{}* Use standard Lean 4 / Mathlib 4 syntax.\par
\smallskip\par
\noindent{}---\par
\noindent{}\#\# lean friendly formalization rules\par
\noindent{}The goal of a formalization LLM is to **choose the library's preferred/canonical formulation** and reduce proof glue, rather than to ``translate literally.''\par
\smallskip\par
\noindent{}1. Prefer the ``canonical API'' (Pick the canonical API)\par
\smallskip\par
\noindent{}**Guideline**: If Mathlib has a canonical API for the same mathematical concept, use it.\par
\noindent{}**Examples**:\par
\smallskip\par
\noindent{}* ``finite support'' \ensuremath{\to} \textasciigrave{}Finsupp.support\textasciigrave{}\par
\noindent{}* ``prime divisors'' \ensuremath{\to} \textasciigrave{}Nat.primeFactors\textasciigrave{} / \textasciigrave{}support\_factorization\textasciigrave{}\par
\noindent{}* ``valuation'' \ensuremath{\to} \textasciigrave{}padicValNat\textasciigrave{} or \textasciigrave{}factorization\textasciigrave{} (pick one; don't mix)\par
\smallskip\par
\noindent{}2. Encode boundary conditions into types when possible (Encode invariants in types)\par
\smallskip\par
\noindent{}**Guideline**: Use stronger types when available (\textasciigrave{}\ensuremath{\mathbb{N}}+\textasciigrave{}, \textasciigrave{}Subsemiring\textasciigrave{}, \textasciigrave{}Units\textasciigrave{}, \textasciigrave{}Subtype\textasciigrave{}) to reduce repeated hypotheses like \textasciigrave{}x \ensuremath{\ne} 0\textasciigrave{}, \textasciigrave{}x \ensuremath{\in} S\textasciigrave{}, etc.\par
\noindent{}**Examples**: positive integers use \textasciigrave{}\ensuremath{\mathbb{N}}+\textasciigrave{}; nonzero elements use \textasciigrave{}\ensuremath{\alpha}\ensuremath{^\times}\textasciigrave{} or \textasciigrave{}Subtype (\ensuremath{\cdot} \ensuremath{\ne} 0)\textasciigrave{}.\par
\smallskip\par
\noindent{}3. Avoid mixing representations: stay within one layer (One representation per theorem)\par
\smallskip\par
\noindent{}**Guideline**: Try to keep a statement within a single ``world'':\par
\smallskip\par
\noindent{}* use only \textasciigrave{}Finsupp.prod\textasciigrave{} or only \textasciigrave{}Finset.prod\textasciigrave{}\par
\noindent{}* use only \textasciigrave{}Set\textasciigrave{} or only \textasciigrave{}Finset\textasciigrave{}\par
\smallskip\par
\noindent{}If you must cross worlds, it's usually best to split into two theorems: a bridging lemma and a main lemma.\par
\smallskip\par
\noindent{}4. Align equation direction with library lemmas (Align direction with lemmas)\par
\smallskip\par
\noindent{}**Guideline**: If the library usually states a lemma as \textasciigrave{}lhs = rhs\textasciigrave{}, write your statement in that direction to avoid scattering \textasciigrave{}.symm\textasciigrave{}.\par
\noindent{}(Direction doesn't change the math, but it affects proof friction.)\par
\smallskip\par
\noindent{}5. Separate explicit constructions from the main statement (Two-tier theorem strategy)\par
\smallskip\par
\noindent{}**Guideline**: By default, output an existence/property theorem; put an explicit witness either in a separate theorem or inside the proof via \textasciigrave{}let\textasciigrave{}.\par
\noindent{}Applies to: situations involving complicated \textasciigrave{}\ensuremath{\prod}/\ensuremath{\sum}\textasciigrave{} expansions, reindexing, or casts.\par
\smallskip\par
\noindent{}6. Expected handling of Classical/DecidableEq\par
\smallskip\par
\noindent{}**Guideline**: If the statement involves \textasciigrave{}Finset\textasciigrave{}, \textasciigrave{}support\textasciigrave{}, \textasciigrave{}toFinset\textasciigrave{}, \textasciigrave{}choose\textasciigrave{}, \textasciigrave{}filter\textasciigrave{}, etc., default to adding \textasciigrave{}by classical\textasciigrave{} in the proof skeleton; don't force \textasciigrave{}classical\textasciigrave{} into the statement itself.\par
\smallskip\par
\noindent{}7. Make ``obviously finite/nonempty'' explicit via concrete structures\par
\smallskip\par
\noindent{}**Guideline**:\par
\smallskip\par
\noindent{}* ``finite'' \ensuremath{\to} use \textasciigrave{}Finset\textasciigrave{} or \textasciigrave{}Fintype\textasciigrave{}\par
\noindent{}* ``nonempty'' \ensuremath{\to} use \textasciigrave{}s.Nonempty\textasciigrave{} or \textasciigrave{}$\exists$ x, x \ensuremath{\in} s\textasciigrave{}\par
\noindent{}* ``only finitely many terms are nonzero'' \ensuremath{\to} use \textasciigrave{}Finsupp\textasciigrave{} or \textasciigrave{}Summable\textasciigrave{}/\textasciigrave{}HasSum\textasciigrave{} (in analysis)\par
\smallskip\par
\smallskip\par
\noindent{}\#\# Skeleton You Must Fill \par
\smallskip\par
\noindent{}You will be given content like:\par
\smallskip\par
\noindent{}"""\par
\noindent{}Please autoformalize the following natural language problem proof step in Lean 4.\par
\noindent{}Use the following lemma name: \{lemma\textbackslash{}\_header\}\par
\noindent{}The natural language statement is: \{statement\}\par
\noindent{}The dependencies are: \{dependencies\}\par
\smallskip\par
\noindent{}This is the lean code skeleton you need to use:\par
\smallskip\par
\noindent{}\textasciigrave{}\textasciigrave{}\textasciigrave{}lean4\par
\noindent{}import Mathlib\par
\noindent{}import Aesop\par
\smallskip\par
\noindent{}set\_option maxHeartbeats 0\par
\smallskip\par
\noindent{}open BigOperators Real Nat Topology Rat Filter\par
\smallskip\par
\noindent{}\{lemma\_header\}\par
\noindent{}[place correct hypothesis here] :\par
\noindent{}[place goal here] := by\par
\noindent{}sorry\par
\noindent{}\textasciigrave{}\textasciigrave{}\textasciigrave{}\par
\noindent{}"""\par
\smallskip\par
\noindent{}The natural language statement has this format: We assume that \textbackslash{}[desciption of lemma l1] \textbackslash{}[l1], that \textbackslash{}[description of lemma l2] \textbackslash{}[l2], ... . From the previous assumptions, we conclude that \textbackslash{}[description of ts\textbackslash{}\_1] \textbackslash{}[ts\textbackslash{}\_1]. The lemma names are provided in square brackets. In the previous example these are "tc1" and "l1", and "ts\textbackslash{}\_1" is the name of the current lemma.\par
\smallskip\par
\noindent{}Your output must **replace** \textasciigrave{}[place correct hypothesis here]\textasciigrave{} with the ordered parameter/hypothesis list (types first, then variables, then dependencies) and replace \textasciigrave{}[place goal here]\textasciigrave{} with the formalized goal.\par
\smallskip\par
\noindent{}If additional previous context is provided (full Lean code of earlier steps), **use it** to extract relevant **type declarations** and **dependency proposition types**, and **include each dependency** as a named hypothesis. Add extra hypothesis if needed.\par
\smallskip\par
\noindent{}**The final fenced block must consist of the header, then exactly one \textasciigrave{}lemma\_header\textasciigrave{} definition, and nothing else.**\par
\smallskip\par
\noindent{}---\par
\smallskip\par
\noindent{}\#\# Few-Shot Examples\par
\smallskip\par
\noindent{}\#\#\# \ding{51} Example 1 \textemdash{} Absolute Value (preserve given hypotheses; demonstrate ordering)\par
\noindent{}**Input**:  \par
\noindent{}* Lemma Name: \textasciigrave{}ts\_1\textasciigrave{}\par
\noindent{}* Natural Language Statement: "Assuming a \ensuremath{\ge} 0 and  b \ensuremath{\le} 0 [tc\_1] and bs (a + b) = abs a + abs b [l1], we can conclude that \textbar{}a + b\textbar{} \ensuremath{\le} \textbar{}a\textbar{} + \textbar{}b\textbar{} [ts\_1]."\par
\noindent{}* Dependencies: ["tc\_1", "l1"]\par
\noindent{}* Lean4 code of tc\_1 and lemma l1 [ommited here]\par
\smallskip\par
\noindent{}**Correct Output**:\par
\noindent{}\textasciigrave{}\textasciigrave{}\textasciigrave{}lean4\par
\noindent{}import Mathlib\par
\noindent{}import Aesop\par
\smallskip\par
\noindent{}set\_option maxHeartbeats 0\par
\smallskip\par
\noindent{}open BigOperators Real Nat Topology Rat Filter\par
\smallskip\par
\noindent{}lemma ts\_1 /- lemma name should be ts\_1-/ \par
\noindent{}\hspace*{1.0em}(a b : \ensuremath{\mathbb{R}})\par
\noindent{}\hspace*{1.0em}(tc\_1 : (a \ensuremath{\ge} 0 $\land$ b \ensuremath{\le} 0))\par
\noindent{}\hspace*{1.0em}(l1 : abs (a + b) = abs a + abs b):\par
\noindent{}\hspace*{1.0em}\textbar{}a + b\textbar{} \ensuremath{\le} \textbar{}a\textbar{} + \textbar{}b\textbar{} := by\par
\noindent{}sorry\par
\noindent{}\textasciigrave{}\textasciigrave{}\textasciigrave{}\textasciigrave{}\par
\smallskip\par
\noindent{}\#\#\# \ding{55} Wrong Example \textemdash{} Replacing Given Hypotheses (don't do this)\par
\smallskip\par
\noindent{}\textasciigrave{}\textasciigrave{}\textasciigrave{}lean4\par
\noindent{}lemma result\par
\noindent{}\hspace*{1.0em}(a b : \ensuremath{\mathbb{R}})\par
\noindent{}\hspace*{1.0em}(h1 : (a \ensuremath{\ge} 0 $\land$ b \ensuremath{\le} 0))\par
\noindent{}\hspace*{1.0em}(h2 : abs (a + b) = abs a + abs b):\par
\noindent{}\hspace*{1.0em}\textbar{}a + b\textbar{} \ensuremath{\le} \textbar{}a\textbar{} + \textbar{}b\textbar{} := by\par
\noindent{}sorry\par
\noindent{}\textasciigrave{}\textasciigrave{}\textasciigrave{}\par
\smallskip\par
\noindent{}*Why wrong: it replaces the original hypothesis names/statements (h1 and h2) the lemma name.*\par
\smallskip\par
\noindent{}\#\#\# \ding{51} Example 2 \textemdash{} Add Missing Variable Types\par
\smallskip\par
\noindent{}**Input**:  \par
\noindent{}* Lemma Name: \textasciigrave{}l3\textasciigrave{}\par
\noindent{}* Natural Language Statement: "Assuming 2 * d = 6 [l2], we can conclude that \textbar{}a + b\textbar{} \ensuremath{\le} \textbar{}a\textbar{} + \textbar{}b\textbar{} [l3]".\par
\noindent{}* Dependencies: ["l2"]\par
\noindent{}* Lean4 code of lemma l2 [ommited here]\par
\smallskip\par
\smallskip\par
\noindent{}**Correct Output**:\par
\smallskip\par
\noindent{}\textasciigrave{}\textasciigrave{}\textasciigrave{}lean4\par
\noindent{}import Mathlib\par
\noindent{}import Aesop\par
\smallskip\par
\noindent{}set\_option maxHeartbeats 0\par
\smallskip\par
\noindent{}open BigOperators Real Nat Topology Rat Filter\par
\smallskip\par
\noindent{}lemma l3\par
\noindent{}\hspace*{1.0em}(d : \ensuremath{\mathbb{R}})\par
\noindent{}\hspace*{1.0em}(l2 : 2 * d = 6) :\par
\noindent{}\hspace*{1.0em}d = 3 := by\par
\noindent{}sorry\par
\noindent{}\textasciigrave{}\textasciigrave{}\textasciigrave{}\par
\smallskip\par
\noindent{}**Why this is correct**: Added missing variable \textasciigrave{}(d : \ensuremath{\mathbb{R}})\textasciigrave{}.\par
\smallskip\par
\noindent{}\#\#\# \ding{51} Example 3 \textemdash{} Matrices with Dependencies\par
\smallskip\par
\noindent{}**Input**: \par
\noindent{}\hspace*{1.0em}* Lemma Name: \textasciigrave{}l4\textasciigrave{}\par
\noindent{}\hspace*{1.0em}* Natural Language Statement: Assuming det (A * B) = det I [l1], det I = 1 [l2], det (A * B) = det A * det B and [l3], we can conclude that \$\textbackslash{}\textbackslash{}det(A) \textbackslash{}\textbackslash{}cdot \textbackslash{}\textbackslash{}det(B) = 1\$ [l4].\par
\noindent{}\hspace*{1.0em}* Dependencies: ["l1", "l2", "l3", "l4"]\par
\noindent{}\hspace*{1.0em}* Lean4 code of lemmas l1, l2, l3 and l4 [ommited here]\par
\smallskip\par
\noindent{}**Correct Output**:\par
\smallskip\par
\noindent{}\textasciigrave{}\textasciigrave{}\textasciigrave{}lean4\par
\noindent{}import Mathlib\par
\noindent{}import Aesop\par
\smallskip\par
\noindent{}set\_option maxHeartbeats 0\par
\smallskip\par
\noindent{}open BigOperators Real Nat Topology Rat Filter\par
\smallskip\par
\noindent{}lemma l4\par
\noindent{}\hspace*{1.0em}\{n : \ensuremath{\mathbb{N}}\} (A B I : Matrix (Fin n) (Fin n) \ensuremath{\mathbb{R}})\par
\noindent{}\hspace*{1.0em}(l1 : det (A * B) = det I)\par
\noindent{}\hspace*{1.0em}(l2 : det I = 1)\par
\noindent{}\hspace*{1.0em}(l3 : det (A * B) = det A * det B) :\par
\noindent{}\hspace*{1.0em}det A * det B = 1 := by\par
\noindent{}sorry\par
\noindent{}\textasciigrave{}\textasciigrave{}\textasciigrave{}\par
\smallskip\par
\noindent{}*(**Rationale**: The LLM correctly identified all the variables (\textasciigrave{}n\textasciigrave{}, \textasciigrave{}A\textasciigrave{}, \textasciigrave{}B\textasciigrave{}, and \textasciigrave{}I\textasciigrave{}) that appear across the target statement and all three dependencies. It declared their types, placing the implicit argument \textasciigrave{}\{n : \ensuremath{\mathbb{N}}\}\textasciigrave{} first, and then formalized the goal, using the probided lean4 code of the dependencies.)*\par
\smallskip\par
\noindent{}---\par
\smallskip\par
\noindent{}\#\# Final Compliance Checklist\par
\smallskip\par
\noindent{}* [ ] **Ordering**: types \ensuremath{\to} explicit variables \ensuremath{\to} dependency hypotheses\par
\noindent{}* [ ] All provided dependencies included as **named hypotheses**\par
\noindent{}* [ ] Goal precisely formalized from the natural language statement\par
\noindent{}* [ ] **Exactly one lemma/theorem** in the block (no other \textasciigrave{}lemma\textasciigrave{}/\textasciigrave{}theorem\textasciigrave{}/\textasciigrave{}def\textasciigrave{}/\textasciigrave{}example\textasciigrave{}/\textasciigrave{}instance\textasciigrave{})\par
\noindent{}* [ ] **Exactly one** \textasciigrave{}:= by\textasciigrave{} and **exactly one** \textasciigrave{}sorry\textasciigrave{}\par
\noindent{}* [ ] Extra hypotheses (e.g., type or variable declarations) are permitted when needed to make the lemma well-formed, but must appear before dependency hypotheses.\par
\noindent{}* [ ] Only use unqualified names from namespaces that are explicitly opened (BigOperators Real Nat Topology Rat Filter); for every other namespace, always write fully qualified names (e.g. Set.union, Function.injective, etc)\par
 \hspace{2em}\par
\noindent{}Please autoformalize the following natural language problem proof step in Lean 4.\par
\noindent{}The natural language statement is: \{item.statement\}\par
\noindent{}The dependencies are: \{dependencies\}\par

\smallskip

\noindent{}**IMPORTANT: Do NOT include import statements or header in your output!**\par
\noindent{}The following header will be automatically added by the system:\par
\noindent{}\textasciigrave{}\textasciigrave{}\textasciigrave{}\par
\noindent{}\{LEAN\_HEADER\}\par
\noindent{}\textasciigrave{}\textasciigrave{}\textasciigrave{}\par

\smallskip

\noindent{}You should ONLY output the lemma/theorem body. Use this skeleton:\par

\smallskip

\noindent{}\textasciigrave{}\textasciigrave{}\textasciigrave{}lean4\par
\noindent{}[place correct hypothesis here] :\par
\noindent{}[place goal here] := by\par
\noindent{}sorry\par
\noindent{}\textasciigrave{}\textasciigrave{}\textasciigrave{}\par

\smallskip

\noindent{}Important: 1 **Please write only one lemma or theorem**!!\par
\noindent{}2 **All variable declarations must be on a SINGLE line, or each must start with the 'variable' keyword.**\par
\noindent{}3 **Function calls need a space before '(': write 'cos (x)' not 'cos(x)'**\par
\noindent{}4 **The formalized code you generate must be faithful to the natural language semantics, ensuring that every condition of the natural language meaning is captured. **\par
\noindent{}5 Be sure to preserve the original meaning of the natural language. It is **not required** that the generated Lean code can be proved, because the natural-language theorem may be unprovable.\par
\endgroup

\end{tcolorbox}
\subsection{Prover}
\label{app:prover}
\begin{tcolorbox}[proverstyle]
\begingroup
\normalfont\normalsize
\raggedright
\sloppy

\noindent{}You are an expert Lean 4 theorem prover.\par
\noindent{}Your job is to complete partially written Lean 4 code and provide correct, verifiable proofs.\par

\smallskip

\noindent{}When given a lemma/theorem and its partial Lean code, you should:\par
\noindent{}- Before producing the Lean 4 code to formally prove the given theorem, provide a detailed proof plan outlining the main proof steps and strategies.\par
\noindent{}- The plan should highlight key ideas, intermediate lemmas, and proof structures that will guide the construction of the final formal proof.\par
\noindent{}- You may add trivial hypotheses (e.g. type declarations) if necessary.\par
\noindent{}- Never remove or rename existing hypotheses; only adapt them in minor ways that preserve meaning.\par

\smallskip

\noindent{}**leanproof tips:**\par
\noindent{}1. Avoid proving ``glue goals'': if you hit goals like False or \_ \ensuremath{\ne} 0 just to feed later steps, prefer changing the quantification domain or using by\_cases instead of hand-deriving facts from membership in support.\par
\noindent{}2. rw matches shapes, not meanings: when rewriting fails, check for representation mismatches (e.g., Finset.prod vs Finsupp.prod) and first align forms via simpa [definition-expansions] using lemma.\par
\noindent{}3. Fix direction with symm first: if you proved the reverse equality, use symm/.symm immediately rather than trying to flip it through rw/simp.\par

\smallskip

\noindent{}- Respond only with Lean 4 code inside a fenced block starting with \textasciigrave{}\textasciigrave{}\textasciigrave{}lean4 and ending with \textasciigrave{}\textasciigrave{}\textasciigrave{}.\par
\noindent{}- Make sure there is no ``sorry'' on the Lean code.\par
\hspace{2em} \par
\noindent{}This is the lemma/theorem I want you to prove:\par
\noindent{}\{item.statement\}\par

\smallskip

\noindent{}\{optional\_lean\_hint\_section\}Complete the following Lean 4 code (**do not remove imports**):\par

\smallskip

\noindent{}\textasciigrave{}\textasciigrave{}\textasciigrave{}lean4\par
\noindent{}\{item.formalization["lean\_code"]\}\par
\noindent{}\textasciigrave{}\textasciigrave{}\textasciigrave{}\par

\smallskip

\noindent{}CRITICAL: You MUST respond with a code block in this exact format:\par
\noindent{}\textasciigrave{}\textasciigrave{}\textasciigrave{}lean4\par
\noindent{}[Your complete Lean 4 proof code here]\par
\noindent{}\textasciigrave{}\textasciigrave{}\textasciigrave{}\par

\smallskip

\noindent{}Important rules:\par
\noindent{}1. **Function calls need a space before '(': write 'cos (x)' not 'cos(x)'**\par
\noindent{}2. Replace 'sorry' with actual proof code.\par
\noindent{}3. Your response MUST include the \textasciigrave{}\textasciigrave{}\textasciigrave{}lean4 code block - this is mandatory.\par
\noindent{}\textasciigrave{}\textasciigrave{}\textasciigrave{}\textasciigrave{}\textasciigrave{}\textasciigrave{}\par

\endgroup
\end{tcolorbox}
\subsection{GEPA Judge}
\label{app:gepa_judge}
\begin{tcolorbox}[gepajudgestyle]
\begingroup
\normalfont\normalsize
\raggedright
\sloppy

\noindent{}Your job is not to run Lean. Decide whether the current proof-graph script is already good enough to enter the actual Lean formalize/prove flow, or whether another statement-level self-reflection rewrite is worth spending one GEPA optimization round.\par

\smallskip

\noindent{}\#\#\# Stage\par
\noindent{}\{stage\}\par

\smallskip

\noindent{}\#\#\# Remaining GEPA optimization rounds\par
\noindent{}\{budget\_remaining\}\par

\smallskip

\noindent{}\#\#\# Original theorem\par
\noindent{}\{nl\_theorem\}\par

\smallskip

\noindent{}\#\#\# Original proof\par
\noindent{}\{nl\_proof\}\par

\smallskip

\noindent{}\#\#\# Target node ids\par
\noindent{}\{json.dumps(sorted(target\_node\_ids), ensure\_ascii=False)\}\par

\smallskip

\noindent{}\#\#\# Current proof graph nodes\par
\noindent{}\textasciigrave{}\textasciigrave{}\textasciigrave{}json\par
\noindent{}\{json.dumps(graph\_payload, ensure\_ascii=False, indent=2)\}\par
\noindent{}\textasciigrave{}\textasciigrave{}\textasciigrave{}\par
\noindent{}\{lean\_feedback\_section\}\par

\smallskip

\noindent{}\#\#\# Criteria\par
\noindent{}1. Evaluate only target nodes. Locked nodes are context only.\par
\noindent{}2. Score every target node against three criteria. Each criterion is worth 1 point, so each node has a maximum score of 3. In \textasciigrave{}reason\textasciigrave{}, give the concrete deduction reasons; if the node gets full marks, the reason may be brief.\par
\noindent{}2.1 Faithfulness to the original proof: the node statement should preserve the semantics of the corresponding step in the original proof.\par
\noindent{}2.2 Strict independent provability of each node: a node may omit necessary conditions from the Original proof, such as \textasciigrave{}k > 0\textasciigrave{} or basic constraints like \textasciigrave{}a\textasciigrave{} is an integer. Such omissions can make the proposition unprovable.\par
\noindent{}2.3 Lean friendliness. Check the following requirements:\par

\smallskip

\noindent{}2.3.1. Relevant-domain Quantification\par

\smallskip

\noindent{}**Problem:** Mathematics often says “for all (x)” even though the argument only needs *relevant* (x), such as nonzero terms, actually occurring terms, or points where a function is defined.\par

\smallskip

\noindent{}**Guideline:** By default, quantify over a clearly specified relevant set or predicate, and make that set or predicate explicit.\par

\smallskip

\noindent{}**NL templates:**\par
\noindent{}* “For every (x) in the set (S), …”\par
\noindent{}* “For every (x) such that (P(x)), …”\par

\smallskip

\noindent{}**Typical triggers:** *all, any, every, for all*.\par

\smallskip

\noindent{}**Benefit:** Reduces Lean work where one must repeatedly prove side conditions like \textasciigrave{}P(x)\textasciigrave{} just to apply hypotheses.\par

\smallskip

\noindent{}2.3.2. Make Domain/Boundary Conditions Explicit\par

\smallskip

\noindent{}**Problem:** Mathematical text often implicitly assumes conditions like nonzero, positive, denominator nonzero, or nonempty set.\par

\smallskip

\noindent{}**Guideline:** Each node should explicitly state the needed side conditions: nonzero, positivity, nonemptiness, invertibility, denominator (\textbackslash{}neq 0), function domain assumptions, etc. You need to use the original full natural-language proof to fill in the domains/ranges and conditions for each variable in this node’s natural-language statement.\par

\smallskip

\noindent{}**NL templates:**\par
\noindent{}* “Assume (x \textbackslash{}neq 0).”\par
\noindent{}* “Assume (S) is finite / nonempty.”\par
\noindent{}* “Assume (f) is defined on … and is continuous on …”\par

\smallskip

\noindent{}2.3.3. Interface Stabilization (Fix the Chosen Notation/API Early)\par

\smallskip

\noindent{}**Problem:** The same concept can have multiple library interfaces, such as valuation/factorization/count, norm/abs, or limit/filter.\par

\smallskip

\noindent{}**Guideline:** Early in the graph, explicitly declare which interface/definition is being used, and stick to it consistently throughout downstream nodes.\par

\smallskip

\noindent{}**NL templates:**\par
\noindent{}* “Let (v\_p(n)) denote …”\par
\noindent{}* “Let (\textbackslash{}lvert x\textbackslash{}rvert) denote the norm …”\par
\noindent{}* “Let (\textbackslash{}lim) be taken with respect to the filter …”\par

\smallskip

\noindent{}2.3.4. Equivalence vs. One-way Implication Annotation\par

\smallskip

\noindent{}**Problem:** Mathematical rewrites are often treated as equivalences, but in Lean the direction matters for which lemmas can be applied.\par

\smallskip

\noindent{}**Guideline:** Every rewrite node should explicitly label whether it is an equivalence (“equivalent to”) or a one-way implication (“implies”).\par

\smallskip

\noindent{}**NL templates:**\par
\noindent{}* “This is equivalent to …”\par
\noindent{}* “This implies …”\par

\smallskip

\noindent{}2.3.5. Type/Cast Awareness\par

\smallskip

\noindent{}**Problem:** Implicit coercions between (\textbackslash{}mathbb\{N\}/\textbackslash{}mathbb\{Z\}/\textbackslash{}mathbb\{Q\}/\textbackslash{}mathbb\{R\}) are a major source of friction in Lean.\par

\smallskip

\noindent{}**Guideline:** The graph output should state which number system each object lives in, and when/where coercions or embeddings are intended.\par

\smallskip

\noindent{}**NL template:**\par
\noindent{}* “View (n) as an integer/real via the canonical embedding.”\par

\smallskip

\noindent{}3. Combining the remaining GEPA optimization rounds and the quality of the target statements, decide whether to enter the actual Lean formalize/prove flow. The goal is that all nodes can pass the Lean flow. Set \textasciigrave{}ready\_for\_lean\textasciigrave{} to true when the current script is good enough to proceed.\par

\smallskip

\noindent{}\#\#\# Example Output Format (content should follow the actual analysis)\par
\noindent{}Return one JSON object:\par
\noindent{}\textasciigrave{}\textasciigrave{}\textasciigrave{}json\par
\noindent{}\{\par
\noindent{}\hspace*{1.5em}"target\_node\_scores": [\par
\noindent{}\hspace*{3em}\{\par
\noindent{}\hspace*{4.5em}"node\_id": "l1",\par
\noindent{}\hspace*{4.5em}"reason": "Concrete deduction reasons; brief if this node receives full marks.",\par
\noindent{}\hspace*{4.5em}"score": 3\par
\noindent{}\hspace*{3em}\}\par
\noindent{}\hspace*{1.5em}],\par
\noindent{}\hspace*{1.5em}"decision\_reason": "Why to proceed to Lean now, or why another self-reflection rewrite is needed.",\par
\noindent{}\hspace*{1.5em}"ready\_for\_lean": true\par
\noindent{}\}\par
\noindent{}\textasciigrave{}\textasciigrave{}\textasciigrave{}\par

\endgroup
\end{tcolorbox}
\subsection{GEPA Rewriter}
\label{app:gepa_rewriter}
\begin{tcolorbox}[geparewriterstyle]
\begingroup
\normalfont\normalsize
\raggedright
\sloppy

\noindent{}You are rewriting proof-decomposition node natural-language statements for Lean formalization.\par

\smallskip

\noindent{}\#\#\# Stage\par
\noindent{}\{stage\}\par

\smallskip

\noindent{}\#\#\# Original theorem\par
\noindent{}\{nl\_theorem\}\par

\smallskip

\noindent{}\#\#\# Original proof\par
\noindent{}\{nl\_proof\}\par

\smallskip

\noindent{}\#\#\# Target node ids to rewrite\par
\noindent{}\{json.dumps(sorted(target\_node\_ids), ensure\_ascii=False)\}\par

\smallskip

\noindent{}\#\#\# Current proof decomposition nodes\par
\noindent{}\textasciigrave{}\textasciigrave{}\textasciigrave{}json\par
\noindent{}\{json.dumps(graph\_payload, ensure\_ascii=False, indent=2)\}\par
\noindent{}\textasciigrave{}\textasciigrave{}\textasciigrave{}\par

\smallskip

\noindent{}\#\#\# Judge feedback\par
\noindent{}\textasciigrave{}\textasciigrave{}\textasciigrave{}json\par
\noindent{}\{json.dumps(judge\_feedback, ensure\_ascii=False, indent=2)\}\par
\noindent{}\textasciigrave{}\textasciigrave{}\textasciigrave{}\par

\smallskip

\noindent{}\#\#\# Rewrite rules\par
\noindent{}1. Output updates only for target nodes that should change.\par
\noindent{}2. Never output updates for locked nodes.\par
\noindent{}3. Keep every node id and dependency list unchanged.\par
\noindent{}4. Preserve the exact mathematical meaning of the original proof step. Do not invent missing proof steps or repair an incorrect proof by changing its meaning.\par
\noindent{}5. Make the statement self-contained and Lean-friendly: include necessary assumptions, domains, type information, nonzero/positivity/boundary conditions, and explicit references to dependency ids when needed.\par
\noindent{}6. Keep the same statement style already used in the decomposition: \textasciigrave{}We assume:\textasciigrave{} and \textasciigrave{}Therefore, we conclude:\textasciigrave{} for lemmas/theorem statements, \textasciigrave{}Premise:\textasciigrave{} for theorem conditions, and \textasciigrave{}Definition:\textasciigrave{} for definitions.\par
\noindent{}7. Each node has a maximum score of 3 in the Judge feedback. The three scoring criteria are: faithfulness to the original proof, strict independent provability of each node, and Lean friendliness. Prefer rewriting nodes with deducted points, and use the Judge reasons to decide what to fix.\par

\smallskip

\noindent{}\#\#\# Trick\par
\noindent{}1. The node’s natural language may omit some necessary conditions from the **original full natural-language context** (e.g., missing k > 0, or basic constraints like a is an integer). If such conditions are necessary for this node, you must carefully identify and include them.\par

\smallskip

\noindent{}2. Enhance the natural-language statement to be more Lean-friendly: make implicit “logical bridges or cindition” explicit, while staying faithful to the semantics of the original natural-language meaning. For example:\par
\noindent{}------\par

\smallskip

\noindent{}2.1. Relevant-domain Quantification\par

\smallskip

\noindent{}**Problem:** Mathematics often says “for all (x)” even though the argument only needs *relevant* (x) (nonzero terms, occurring terms, points where a function is defined, etc.).\par
\noindent{}**Guideline:** By default, quantify over a clearly specified relevant set or predicate, and make that set/predicate explicit.\par
\noindent{}**NL templates:**\par

\smallskip

\noindent{}* “For every (x) in the set (S), …”\par
\noindent{}* “For every (x) such that (P(x)), …”\par

\smallskip

\noindent{}**Typical triggers:** *all, any, every, for all*.\par
\noindent{}**Benefit:** Reduces Lean “glue” work where you must repeatedly prove side conditions like \textasciigrave{}P(x)\textasciigrave{} just to apply hypotheses.\par

\smallskip

\noindent{}2.2. Make Domain/Boundary Conditions Explicit\par

\smallskip

\noindent{}**Problem:** Mathematics often implicitly assumes conditions like “nonzero/positive/denominator nonzero/nonempty set.”\par
\noindent{}**Guideline:** Each node should explicitly state the needed side conditions: nonzero, positivity, nonemptiness, invertibility, denominator (\textbackslash{}neq 0), function domain assumptions, etc. **You need to use the original full natural-language proof to fill in the domains/ranges and conditions for each variable in this node’s natural-language statement!!!!**\par
\noindent{}**NL templates:**\par

\smallskip

\noindent{}* “Assume (x \textbackslash{}neq 0).”\par
\noindent{}* “Assume (S) is finite / nonempty.”\par
\noindent{}* “Assume (f) is defined on … and is continuous on …”\par

\smallskip

\noindent{}2.3. Interface Stabilization (Fix the Chosen Notation/API Early)\par

\smallskip

\noindent{}**Problem:** The same concept can have multiple library interfaces (e.g., valuation/factorization/count; norm/abs; limit/filter).\par
\noindent{}**Guideline:** Early in the graph, explicitly declare which interface/definition is being used, and stick to it consistently throughout downstream nodes.\par
\noindent{}**NL templates:**\par

\smallskip

\noindent{}* “Let (v\_p(n)) denote …”\par
\noindent{}* “Let (\textbackslash{}lvert x\textbackslash{}rvert) denote the norm …”\par
\noindent{}* “Let (\textbackslash{}lim) be taken with respect to the filter …”\par

\smallskip

\noindent{}2.4. Equivalence vs. One-way Implication Annotation\par

\smallskip

\noindent{}**Problem:** Mathematical rewrites are often intended as equivalences, but in Lean the *direction* matters for which lemmas can be applied.\par
\noindent{}**Guideline:** Every rewrite node should explicitly label whether it is an equivalence (“equivalent to”) or a one-way implication (“implies”).\par
\noindent{}**NL templates:**\par

\smallskip

\noindent{}* “This is equivalent to …”\par
\noindent{}* “This implies …”\par

\smallskip

\noindent{}2.5. Type/Cast Awareness\par

\smallskip

\noindent{}**Problem:** Implicit coercions between (\textbackslash{}mathbb\{N\}/\textbackslash{}mathbb\{Z\}/\textbackslash{}mathbb\{Q\}/\textbackslash{}mathbb\{R\}) are a major source of friction in Lean.\par
\noindent{}**Guideline:** The graph output should state which number system each object lives in, and when/where coercions (embeddings) are intended.\par
\noindent{}**NL template:**\par

\smallskip

\noindent{}* “View (n) as an integer/real via the canonical embedding.”\par

\smallskip

\noindent{}\#\#\# Output\par
\noindent{}Return one JSON object:\par
\noindent{}\textasciigrave{}\textasciigrave{}\textasciigrave{}json\par
\noindent{}\{\par
\noindent{}\hspace*{1.5em}"updates": [\par
\noindent{}\hspace*{3em}\{\par
\noindent{}\hspace*{4.5em}"node\_id": "l1",\par
\noindent{}\hspace*{4.5em}"natural\_language": "Replacement natural language for this node.",\par
\noindent{}\hspace*{4.5em}"statement": "Replacement self-contained statement for this node.",\par
\noindent{}\hspace*{4.5em}"lean\_hint": "Optional short Lean hint."\par
\noindent{}\hspace*{3em}\}\par
\noindent{}\hspace*{1.5em}]\par
\noindent{}\}\par
\noindent{}\textasciigrave{}\textasciigrave{}\textasciigrave{}\par

\endgroup
\end{tcolorbox}
\subsection{Evaluation Criteria}
\label{app:eval}

\begin{tcolorbox}[evalstyle, breakable]
\#\# Guidelines for Consistency Check

Core Checking Requirements:

- When a critique from a previous autoformalization and consistency check result is provided, you must first analyze its findings and then assess their problems.

- Must carefully compare the Natural Language and Lean code through a rigorous and explicit process.

- Determine if the Lean theorem statement is an exact and faithful formalization of the mathematical problem and proof.

- If any result is Incorrect of consistency, briefly list all inconsistencies and reasons leading to the Incorrect determination in comments.

\#\# Evaluation Stages (All required):

1. Math Assertion and Proof Analysis

Identify all structurally and semantically relevant components of the mathematical problem and proof, including variables, types, quantifiers, constraints, logic structure, conclusion, and so on. The analysis should be based on the actual content of the text.

2. Lean Statement Analysis

Extract all structurally and semantically relevant components from the Lean statement, including variables, types, conditions, quantifiers, constraints, the final claim, and so on. The analysis should reflect the actual content present in the Lean code.

3. Comparative Verification

Check for exact correspondence between the math and Lean statements; you may refer to aspects like:

- Semantic alignment, proof-step alignment, logic structure, and quantifier correctness.

- Preservation of constraints and boundary assumptions.

- Accurate typing and use of variables.

- Syntactic validity and proper Lean usage (free from errors).

- Use of symbols and constructs without semantic drift.

- No missing elements, no unjustified additions, and no automatic corrections or completions.

4. Final Judgement

Based solely on the above analysis, judge whether the Lean statement is a correct and exact formalization of the mathematical problem and proof.

5. Accuracy Confirmation

If correct: clearly confirm why all elements match.

If incorrect: list all mismatches and explain how each one affects correctness.

\end{tcolorbox}
\newpage
\section{Model Configuration}

\label{app:model_configuration}
\begin{table}[H]
\centering
\caption{Models used by different methods.}
\label{tab:models}

\renewcommand{\arraystretch}{1.35}
\setlength{\tabcolsep}{5pt}

\setlength{\lightrulewidth}{0.25pt}

\small

\begin{tabularx}{\textwidth}{%
>{\raggedright\arraybackslash}p{0.24\textwidth}
>{\raggedright\arraybackslash}p{0.35\textwidth}
>{\raggedright\arraybackslash}p{0.33\textwidth}
}
\toprule
\textbf{Method} 
& \textbf{Formal Language Model} 
& \textbf{Natural Language Model} \\
\midrule

\textbf{Monotonic}
&
DeepSeek-Prover-V2-7B~\citep{ren2025deepseek}\newline
Goedel-Prover-V2-8B~\citep{lin2025goedel}
&
Seed-Coder-8B~\citep{seedcoder}\newline
Qwen2.5-7B~\citep{qwen25}\newline
GPT-4.1-mini~\citep{openai2024gpt4technicalreport}
\\
\midrule

\textbf{ProofBridge}
&
Kimina-Prover-RL-1.7B fine-tuned with SFT~\citep{wang2025kimina}\newline
NL encoder initialized from all-MiniLM-L6-v2
&
None
\\
\midrule

\textbf{ProofFlow}
&
Goedel-Formalizer-V2-32B\newline
Goedel-Prover-V2-32B
&
Gemini3-Pro~\citep{team2023gemini}
\\
\midrule

\textbf{\ToMap{} w/ Qwen3-30B-A3B-Instruct-2507}
&
Goedel-Formalizer-V2-32B\newline
Goedel-Prover-V2-32B
&
Qwen3-30B-A3B-Instruct-2507~\citep{yang2025qwen3}
\\
\midrule

\textbf{\ToMap{} w/ Gemini3-Pro}
&
Goedel-Formalizer-V2-32B\newline
Goedel-Prover-V2-32B
&
Gemini3-Pro
\\

\bottomrule
\end{tabularx}
\end{table}

\section{Hyperparameter Settings}
\label{app:Hyperparameter_Settings}
\begin{table}[H]
\centering
\caption{Hyperparameters used in \ToMap{}.}
\renewcommand{\arraystretch}{1.25}
\begin{tabular*}{0.7\textwidth}{@{\extracolsep{\fill}}lll@{\hspace{1.5em}}}
\toprule
\multicolumn{2}{c}{\textbf{Hyperparameter}} & \textbf{Value} \\
\midrule
\multirow{2}{*}{Decomposer} 
    & Context length & 32000 \\
    & Temperature & 0.7 \\
\addlinespace[0.4em]

\multirow{2}{*}{Formalizer} 
    & Context length & 12000 \\
    & Temperature & 0.7 \\
\addlinespace[0.4em]

\multirow{2}{*}{Prover} 
    & Context length & 12000 \\
    & Temperature & 0.7 \\
\addlinespace[0.4em]

\multirow{2}{*}{GEPA Judge} 
    & Context length & 32000 \\
    & Temperature & 0.7 \\
\addlinespace[0.4em]

\multirow{2}{*}{GEPA Rewriter} 
    & Context length & 32000 \\
    & Temperature & 0.7 \\
\addlinespace[0.4em]

GEPA\_MAX\_ITERS & & 10 \\
\addlinespace[0.4em]

Gate threshold $\epsilon$ & & 1 \\
\bottomrule
\end{tabular*}
\label{tab:hyperparameters}
\end{table}

\section{Decomposition Optimization Examples}
\label{app:Decomposition_Optimization_Examples}
This section presents several examples of decomposition optimization, illustrating that optimizing decompositions is of practical significance for preserving the semantics of the original proof and for subsequently generating correct Lean code. The optimization specifically includes aspects such as eliminating ambiguity in natural language, uncovering implicit assumptions, and improving the definitions of variables.
\subsection{Example 1}
Some conventional notations in natural language are ambiguous. For example, (i) is usually understood by default as the imaginary unit in the complex numbers, whereas in Lean, (i) must be rigorously defined.

\textbf{Natural language subclaim before optimization.}

We assume:
\begin{itemize}
    \item $x,y \in \mathbb{R}$;
    \item $z=x+iy$.
\end{itemize}

Therefore, we conclude:
\begin{itemize}
    \item $(1+i)z=(x-y)+i(x+y)$ .
\end{itemize}

\textbf{Lean subclaim before optimization.}

\noindent\texttt{import Mathlib}\\
\texttt{import Aesop}\\

\noindent\texttt{set\_option maxHeartbeats 0}\\

\noindent\texttt{open BigOperators Real Nat Topology Rat Filter}\\

\noindent\texttt{theorem l2 (x y : }$\mathbb{R}$\texttt{) (z : }$\mathbb{C}$\texttt{) (h : z = x + I * y) :}\\
\hspace*{2em}\texttt{(1 + I) * z = (x - y) + I * (x + y) := by sorry}

\textbf{Natural language subclaim after optimization.}

We assume:
\begin{itemize}
    \item $x,y \in \mathbb{R}$ ;
    \item $z=x+iy$, where $i$ is the standard imaginary unit in $\mathbb{C}$ with $i^2=-1$.
\end{itemize}

Therefore, we conclude:
\begin{itemize}
    \item $(1+i)z=(x-y)+i(x+y)$ as an equality in $\mathbb{C}$.
\end{itemize}

\textbf{Lean subclaim after optimization.}

\noindent\texttt{import Mathlib}\\
\texttt{import Aesop}\\

\noindent\texttt{set\_option maxHeartbeats 0}\\

\noindent\texttt{open BigOperators Real Nat Topology Rat Filter}\\

\noindent\texttt{theorem l2}\\
\hspace*{2em}\texttt{(x y : }$\mathbb{R}$\texttt{)}\\
\hspace*{2em}\texttt{(z : }$\mathbb{C}$\texttt{)}\\
\hspace*{2em}\texttt{(h1 : z = x + y * Complex.I) :}\\
\hspace*{2em}\texttt{(1 + Complex.I) * z = (x - y) + (x + y) * Complex.I := by sorry}
\subsection{Example 2}
The decomposition fails to capture an implicit condition in the original proof: in the original proof, $(a,b)$ is nonempty, i.e., $a < b$, whereas the decomposition imposes no such restriction.

\textbf{Natural language theorem.}

Let $a,b \in \mathbb{Q}$. Then for any integer $n$,
\[
\#\left((a,b)\cap \mathbb{Z}\right)
\equiv
\#\left((a,b+2n)\cap \mathbb{Z}\right)
\pmod{2}
\]
and
\[
\#\left((a,b)\cap \mathbb{Z}\right)
\equiv
\#\left((a-2n,b)\cap \mathbb{Z}\right)
\pmod{2},
\]
provided that each interval involved in the congruence is nonempty.

\textbf{Natural language proof.}

Let $\lceil x\rceil$ denote the least integer greater than or equal to $x$.

Since $n>0$,
\[
(a,b+2n)=(a,b)\cup [b,b+2n),
\]
where the union is disjoint. There are $2n$ integers
\[
\lceil b\rceil,\ \lceil b\rceil+1,\ \ldots,\ \lceil b\rceil+2n-1
\]
in the interval $[b,b+2n)$, so the first congruence of the lemma is true in this case.

We also have
\[
(a,b-2n)=(a,b)\text{ minus }[b-2n,b),
\]
and $[b-2n,b)$ contains exactly $2n$ integers, so the lemma is also true when $n$ is negative.

The statement about
\[
\#\left((a-2n,b)\cap \mathbb{Z}\right)
\]
is proved in a similar manner.

\textbf{Natural language subclaim before optimization.}

We assume:
\begin{itemize}
    \item $a,b \in \mathbb{Q}$ ;
    \item $n \in \mathbb{Z}$ with $n>0$.
\end{itemize}

Therefore, we conclude:
\begin{itemize}
    \item $(a,b+2n)=(a,b)\cup [b,b+2n)$.
\end{itemize}

\textbf{Lean subclaim before optimization.}

\noindent\texttt{import Mathlib}\\
\texttt{import Aesop}\\

\noindent\texttt{set\_option maxHeartbeats 0}\\

\noindent\texttt{open BigOperators Real Nat Topology Rat Filter}\\

\noindent\texttt{theorem l1}\\
\hspace*{2em}\texttt{(a b : }$\mathbb{Q}$\texttt{) (n : }$\mathbb{Z}$\texttt{) (hn : n > 0)}\\
\hspace*{2em}\texttt{(tc\_1 : True) :}\\
\hspace*{2em}\texttt{Set.Ioo a (b + 2 * n) = Set.Ioo a b }$\cup$\texttt{ Set.Ico b (b + 2 * n) }$\land$\\
\hspace*{2em}\texttt{Disjoint (Set.Ioo a b) (Set.Ico b (b + 2 * n)) := by}\\
\hspace*{2em}\texttt{sorry}

\textbf{Natural language subclaim after optimization.}

We assume:
\begin{itemize}
    \item $a,b \in \mathbb{Q}$;
    \item $n \in \mathbb{Z}$;
    \item $n>0$;
    \item $a<b$ as real numbers, since the interval is nonempty.
\end{itemize}

Therefore, we conclude:
\begin{itemize}
    \item As sets of real numbers,
    \[
    (a,b+2n)=(a,b)\cup [b,b+2n)
    \]
\end{itemize}

\textbf{Lean subclaim after optimization.}

\noindent\texttt{import Mathlib}\\
\texttt{import Aesop}\\

\noindent\texttt{set\_option maxHeartbeats 0}\\

\noindent\texttt{open BigOperators Real Nat Topology Rat Filter}\\

\noindent\texttt{theorem l1}\\
\hspace*{2em}\texttt{(a b : }$\mathbb{Q}$\texttt{) (n : }$\mathbb{Z}$\texttt{)}\\
\hspace*{2em}\texttt{(hn : n }$>$\texttt{ 0) (hab : a }$<$\texttt{ b) :}\\
\hspace*{2em}\texttt{(Set.Ioo (a : }$\mathbb{R}$\texttt{) (b + 2 * n : }$\mathbb{R}$\texttt{) =}\\
\hspace*{4em}\texttt{Set.Ioo (a : }$\mathbb{R}$\texttt{) (b : }$\mathbb{R}$\texttt{) }$\cup$\\
\hspace*{4em}\texttt{Set.Ico (b : }$\mathbb{R}$\texttt{) (b + 2 * n : }$\mathbb{R}$\texttt{)) }\texttt{:= by}\\
\hspace*{2em}\texttt{sorry}
\subsection{Example 3}
For the Lean language, each variable should be rigorously defined, for example by specifying which number domain it belongs to. Such conditions may exist in the natural language, but may not have been captured in the initial version of a given subclaim.

\textbf{Natural language subclaim before optimization.}

We assume:
\begin{itemize}
    \item $B=30$ ;
    \item $h=6.5$ .
\end{itemize}

Therefore, we conclude:
\begin{itemize}
    \item $\dfrac{1}{3}Bh=65$ .
\end{itemize}

\textbf{Natural language subclaim after optimization.}

We assume:
\begin{itemize}
    \item Let $B,h \in \mathbb{R}$;
    \item $B=30$ ;
    \item $h=6.5$ .
\end{itemize}

Therefore, we conclude:
\begin{itemize}
    \item $\dfrac{1}{3}Bh=65$ .
\end{itemize}

\newpage

\end{document}